\documentclass[conference,10pt]{IEEEtran}
\usepackage{amsmath}
\usepackage{amsthm}
\usepackage{amsfonts}
\usepackage{graphicx}
\usepackage{dsfont}
\usepackage{subcaption}
\usepackage{booktabs}
\usepackage{flushend}
\usepackage{enumerate}
\usepackage[shortlabels]{enumitem}
\usepackage{algorithmic}
\usepackage{graphics}
\usepackage[numbers]{natbib}
\usepackage{graphicx}
\usepackage{multirow}
\usepackage{bigdelim}
\usepackage[table]{xcolor}    
\usepackage{pifont}
\usepackage{amssymb}
\usepackage{url}
\usepackage{hyperref}
\usepackage[capitalise]{cleveref}

\usepackage{newfloat}
\DeclareFloatingEnvironment[
    fileext=los,
    listname={List of Protocols},
    name=Protocol,
    placement=tbhp
]{protocol}

\hypersetup{
    colorlinks,
    linkcolor={red!50!black},
    citecolor={blue!50!black},
    urlcolor={blue!80!black}
}

\crefname{protocol}{Protocol}{Protocols}
\crefname{crafter}{Crafter}{Crafters}
\crefname{distinguisher}{Distinguisher}{Distinguishers}

\usepackage{overpic}
\usepackage{algorithm}
\usepackage{algorithmic}
\newcommand{\paragraphb}[1]{\noindent{\bf #1} }

\usepackage{colortbl}

\usepackage{cryptocode}

\usepackage{bbm}
\newcommand{\alga}{{Crafter}}
\newcommand{\algb}{{Distinguisher}}

\newcommand{\fix}[1]{#1}

\usepackage[utf8]{inputenc}
\usepackage{pgfplots}
\DeclareUnicodeCharacter{2212}{−}
\usepgfplotslibrary{groupplots,dateplot}
\usetikzlibrary{patterns,shapes.arrows}
\tikzstyle{block} = [rectangle, draw, text width=6em, text centered, rounded corners, minimum height=4em]

\pgfplotsset{compat=newest}

\newlength\figureheight
\newlength\figurewidth
\newfloat{crafter}{htbp}{loa}
\floatname{crafter}{Crafter} 

\newfloat{distinguisher}{htbp}{loa}
\floatname{distinguisher}{Distinguisher} 

\begin{document}
\title{Adversary Instantiation: Lower Bounds for Differentially Private Machine Learning}



\author{

	\IEEEauthorblockN{Milad Nasr\IEEEauthorrefmark{1},
	Shuang Song\IEEEauthorrefmark{2},
	Abhradeep  Thakurta\IEEEauthorrefmark{2},
	Nicolas Papernot\IEEEauthorrefmark{2} and
	Nicholas Carlini\IEEEauthorrefmark{2}}
	\IEEEauthorblockA{\IEEEauthorrefmark{1}University of Massachusetts Amherst \hspace{3cm}
	\IEEEauthorrefmark{2}Google Brain}
	\IEEEauthorblockA{\IEEEauthorrefmark{1}milad@cs.umass.edu \hspace{1cm}
	\IEEEauthorrefmark{2}\{shuangsong, athakurta, papernot, ncarlini\}@google.com}
	}
	
\date{}
\maketitle

\graphicspath{{./draftyfigures/},}

\section*{Abstract}
Differentially private (DP) machine learning allows us to train models
on private data while limiting data leakage. 
%
DP formalizes this data leakage through a cryptographic game, where an adversary must predict
if a model was trained on a dataset $D$, or a dataset $D'$ that differs in just one example.
If observing the training algorithm does not meaningfully increase the adversary's odds of successfully guessing which dataset the model was trained on, then the
algorithm is said to be differentially private. Hence, the purpose of privacy analysis is to  upper bound the probability that any adversary could successfully guess which dataset the model was trained on.
%

In our paper, we instantiate this hypothetical adversary
in order to establish lower bounds on  the probability that  this distinguishing
game can be won.
We use this adversary to evaluate the importance of the adversary capabilities
allowed in the privacy analysis of DP training algorithms.
%


For DP-SGD, the most common method for training neural
networks with differential privacy,
our lower bounds are tight and match the theoretical upper bound.
This implies that in order to prove better upper bounds, it will be necessary
to make use of additional assumptions.
Fortunately, we find that our attacks \emph{are} significantly
weaker when additional (realistic)  restrictions are put in place on the adversary's capabilities. 
%
Thus, in the practical setting common to many real-world deployments, there is a gap between our lower bounds and the upper bounds provided by the analysis:  differential privacy is conservative and adversaries may
not be able to leak as much information as suggested by the theoretical bound.


\section{Introduction}

With machine learning now being used to train models on sensitive user data,
ranging from medical images \cite{esteva2017dermatologist}, 
to personal email \cite{smartcompose} and 
text messages \cite{adp2017learning}, it is becoming ever more important that these
models are privacy preserving.
This privacy is both desirable for users, and increasingly often also legally mandated in frameworks such as the GDPR \cite{cohen2020towards}.

Differential privacy (DP) is now the standard definition for privacy \cite{dwork2006calibrating,dwork2008differential}.
While first defined as a property a query mechanism satisfies on a database,
differential privacy analysis has since been extended to algorithms for training machine
learning models on private training data
~\cite{chaudhuri2011differentially,bassily2014private,song2013stochastic,abadi2016deep,mcmahan2017learning, wu2017bolt, iyengar2019towards, pichapati2019adaclip,feldman2020private,feldman2018privacy}.
On neural networks, differentially private stochastic gradient 
descent (DP-SGD) \cite{abadi2016deep,bassily2014private,song2013stochastic}
is the most popular method to train neural networks with DP guarantees.
Differential privacy sets up a game where the adversary is trying to guess whether a training algorithm took as its input one dataset $D$ or a second 
dataset $D'$ that differs in only one example. If observing the training algorithm's outputs allows the adversary to improve their odds of guessing correctly, then the algorithm leaks private information. Differential privacy proposes to randomize the algorithm in such a way that it becomes possible to analytically upper bound the probability of an adversary making a successful guess, hence quantifying the maximum leakage of private information. 

 In recent work~\cite{jagielski2020auditing} proposed to audit the privacy guarantees of DP-SGD by instantiating a relatively weak, black-box adversary who observed the model's predictions.   In this paper, we instantiate this adversary with a spectrum of attacks that spans from a black-box adversary (that is only able to observe the model's predictions) to a worst-case yet often unrealistic adversary (with the ability to poison training 
gradients and observe intermediate model updates during training). This not only enables us to provide a stronger lower bounds on the privacy leakage (compared to~\cite{jagielski2020auditing}), it also helps us identify which capabilities are needed for adversaries to be able to extract exactly as much private information as is possible given the upper bound provided by the DP analysis. 
%

Indeed, in order to provide strong, composable guarantees that avoid the pitfalls of other
privacy analysis methods \cite{li2007t,machanavajjhala2007diversity,sweeney2002k}, DP and DP-SGD assume the existence of 
powerful adversaries that may not be practically realizable.
%

\begin{figure}
    \centering
\begin{tikzpicture}

\definecolor{color0}{rgb}{0.12156862745098,0.466666666666667,0.705882352941177}
\definecolor{color1}{rgb}{1,0.498039215686275,0.0549019607843137}
\definecolor{color2}{rgb}{0.172549019607843,0.627450980392157,0.172549019607843}
\definecolor{color3}{rgb}{0.572549019607843,0.0,0.572549019607843}

\begin{axis}[
height=\figureheight,
width=\figurewidth,
legend cell align={left},
legend style={fill opacity=0.8, draw opacity=1, text opacity=1, at={(0.03,0.97)}, anchor=north west, draw=white!80!black},
tick align=outside,
tick pos=left,
x grid style={white!69.0196078431373!black},
xmajorgrids,
xmin=-0.3, xmax=5.3,
xtick style={color=black},
y grid style={white!69.0196078431373!black},
ymajorgrids,
ymin=0, ymax=2.675,
ytick style={color=black},
xticklabels={,API,Blackbox,Whitebox,Adaptive,Gradient,Dataset},
x tick label style={rotate=45,anchor=east},
ylabel={measured privacy}
]
\draw[draw=none,fill=color0] (axis cs:-0.2,0) rectangle (axis cs:0.2,0.22);

\draw[draw=none,fill=color0] (axis cs:0.8,0) rectangle (axis cs:1.2,1.22);
\draw[draw=none,fill=color0] (axis cs:1.8,0) rectangle (axis cs:2.2,1.32);
\draw[draw=none,fill=color0] (axis cs:2.8,0) rectangle (axis cs:3.2,1.72);
\draw[draw=none,fill=color0] (axis cs:3.8,0) rectangle (axis cs:4.2,1.82);


\draw[draw=none,fill=color0] (axis cs:4.8,0) rectangle (axis cs:5.2,1.98);

\path [draw=black, semithick]
(axis cs:0,0.01)
--(axis cs:0,0.43);

\path [draw=black, semithick]
(axis cs:1,0.221)
--(axis cs:1,2.219);

\path [draw=black, semithick]
(axis cs:2,0.2451)
--(axis cs:2,2.3949);

\path [draw=black, semithick]
(axis cs:3,0.351)
--(axis cs:3,3.089);

\path [draw=black, semithick]
(axis cs:4,0.401)
--(axis cs:4,3.239);











\path [draw=black, semithick]
(axis cs:5,1.9)
--(axis cs:5,2.06);

\addplot [semithick, red, dashed]
table {%
-0.3 2
5.3 2
};
\end{axis}

\end{tikzpicture}
\hspace*{1cm}
\begin{tikzpicture}
\node  (A) {Practical};
\node [right=3cm of A] (B) {Theoretical};
\draw [->] (A) -- (B);
\end{tikzpicture}
    \caption{\textbf{Summary of our results}, plotting emperically measured $\varepsilon$
    when training a model with
    $\varepsilon=2$ differential privacy
    on MNIST.
    The dashed red line corresponds to the certifiable upper bound.
    Each bar correspond to the privacy offered by
    increasingly powerful adversaries.
    In the most realistic setting, training with privacy offers much more
    empirically measured privacy.
    When we provide full attack capabilities,
    our lower bound shows that the DP-SGD upper bound
    is tight.
    }
    \label{fig:main}
\end{figure}
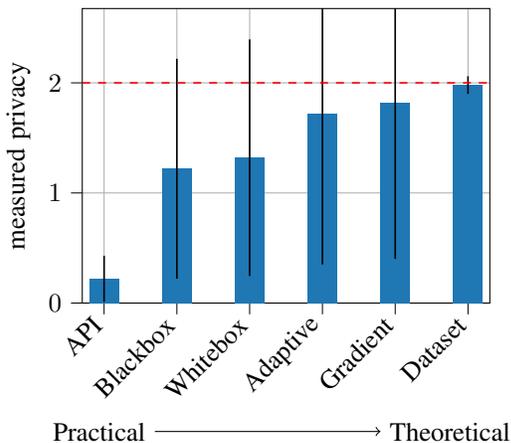
For example, the DP-SGD analysis assumes that
the intermediate computations used
to train a model are published to the adversary,
when in most practical settings only the
final, fully-trained model is revealed.
This is done not because it is desirable---but because there are limited (known) ways to 
improve the analysis by assuming the adversary has access to only one model~\cite{feldman2018privacy}.
As such, it is conceivable---and indeed likely---that
the actual privacy in practical scenarios is greater than
what can be shown theoretically (indeed, DP-SGD is known to be  tight asymptotically~\cite{bassily2014private}).

Researchers have often assumed that this will be true \cite{jayaraman2019evaluating},
and argue that it is acceptable to train models with guarantees
so weak that they are essentially vacuous, hoping that the
actual privacy offered is much stronger \cite{carlini2019secret}. 
This is important because training models with weak guarantees generally allows one to achieve greater model utility, which often leads practitioners to  follow this train of thought when tuning their DP training algorithms.

\subsection{Our Techniques}
We build on the poisoning adversary introduced in Jagielski \emph{et al.} \cite{jagielski2020auditing}. By adversarially constructing two datasets $D$ and $D'$ we can \emph{audit} DP-SGD.
Concretely, we instantiate the hypothetical differential privacy adversary
under various adversary scenarios
by providing a pair of concrete attack algorithms: one that constructs the two datasets $D$ and $D'$
differing in one example, and another that receives a model as input (that was trained on
one of these two datasets) and predicts which one it was trained on.

We use these two adversaries as a tool to measure the relative importance of the
assumptions made by the DP-SGD analysis, as well as the potential
benefits of assumptions that are not currently made, but could be reasonable
assumptions to make in the future.
Different combinations of assumptions correspond to different threat models and constraints on the adversary's capabilities, as exposed in Sections~\ref{sec:attack1} to \ref{sec:attack6}.

There are two communities who this analysis impacts.
\begin{itemize}
    \item Practitioners would like to know the privacy leakage in situations that are as close to realistic as possible.
    Even if it is not possible to prove tight upper
    bounds of privacy, we can provide best-effort empirical evidence that the privacy offered
    by DP-SGD is greater than what is proven when the adversary is more constrained.

    \item Theoreticians, conversely, care about identifying ways to improve the
    current analysis.
    By studying various capabilities that a real-world adversary would actually
    have, but that the analysis of DP-SGD does not assume are placed on the
    adversary, we are able to estimate the potential utility of introducing
    additional assumptions.
\end{itemize}

\subsection{Our Results}

Figure~\ref{fig:main} summarizes our key results.
When our adversary is given full capabilities, our lower bound
matches the provable upper bound, demonstrating \textbf{the DP-SGD analysis is tight in the worst-case}.
In the past, more sophisticated analysis techniques (without making new assumptions) 
were able to establish better and
better upper bounds on essentially the same DP-SGD algorithm \cite{abadi2016deep}.
This meant that we could train a model once, and over time its privacy would ``improve''  (so to speak),
as progress in theoretical work on differential privacy yielded a  tighter  analysis of the same algorithm used to train the model. 
Our lower bound implies that this trend has come to an end, and new
assumptions will be necessary to establish tighter bounds.

Conversely, \textbf{the bound \fix{may be} loose under realistic settings}, such as 
when we assume a more restricted adversary that is only allowed to view the
final trained model (and not every intermediate model obtained during training). In this setting, 
our bound is substantially lowered
\fix{and as a result it is possible that better analysis might be able
to improve the privacy bound.}

Finally, we show that many of the
capabilities allowed by the adversary
do not significantly strengthen the attack.
For example, the DP-SGD analysis assumes the adversary is given
access to all intermediate models generated throughout training;
surprisingly, we find that access to \emph{just} the final model
is almost as good as \emph{all} prior models.

On the whole, our results have broad implications for those deploying differentially
private machine learning in practice (for example, indicating situations where the empirical
privacy is likely stronger than the worst case) and also implications for
those theoretically analyzing properties of DP-SGD (for example, indicating that
new assumptions will be required to obtain stronger privacy guarantees).

\section{Background \& Related Work}

\subsection{Machine Learning} 

A machine learning model is a parameterized function $f_\theta \colon \mathcal{X} \to \mathcal{Y}$
that takes inputs from an input space $\mathcal{X}$ and returns outputs from an output space $\mathcal{Y}$.
The majority of this paper focuses on the class of functions $f_\theta$ known as 
deep neural networks \cite{lecun1998gradient}, represented as a sequence of linear layers with 
non-linear activation functions \cite{nair2010rectified}.
Our results are independent of any details of the neural network's architecture.

The model parameters $\theta$ are obtained through a \emph{training algorithm}
$\mathcal{T}$
that minimize the average loss
$\ell(f_\theta, x, y)$ 
on a finite-sized \emph{training datasets}
$D = \{(x_i, y_i)\}_{i=1}^{|D|}$,
denoted by $\mathcal{L}(f_\theta, D)$.
Formally, we write $f_\theta \gets \mathcal{T}(D).$

Stochastic gradient descent \cite{lecun1998gradient} is the canonical method for minimizing this
loss.
We sample a mini-batch of examples 
\[\mathbb{B}(D) = B = \{(x_i, y_i)\}_{i=1}^{|B|} \subset D.\]

Then, we compute the average mini-batch loss as
\[\mathcal{L}(f_{\theta}; B) = {1 \over |B|}\sum_{(x,y) \in B}\ell(f_{\theta}; x, y)\]
and update the model parameters according to
\begin{equation}
    \label{eqn:sgd}
    \theta_{i+1} \gets \theta_i - \eta \nabla_{\theta}\mathcal{L}(f_{\theta_{i}}; B)
\end{equation} 
taking a step of size $\eta$, the \emph{learning rate}.
We abbreviate each step of this update rule as $f_{\theta+1} \gets \mathcal{S}(f_\theta, B)$.

Because neural networks are trained on a finite-sized training dataset,
models train for multiple \emph{epochs} repeating the same
examples over and over, often tens to hundreds of times.
This has consequences for the privacy of the training data.
Machine learning models are often significantly over-parameterized
\cite{song2017machine,zhang2016understanding}: they contain sufficient
capacity to memorize the particular aspects of the data they are trained on, even if these aspects are irrelevant to final accuracy.
This allows a wide range of attacks that leak information about the training data
given access to the trained model.
These attacks range from membership inference attacks~\cite{carlini2019secret,hayes2019logan,melis2019exploiting,nasr2019comprehensive,song2017machine,shokri2017membership} to training data extraction
attacks~\cite{fredrikson2015model,melis2019exploiting}.

\subsection{Differential Privacy}
Differential privacy (DP)~\cite{dwork2006calibrating,dwork2006our,dwork2014algorithmic} has become the de-facto definition of algorithmic privacy.
An algorithm $\mathcal{M}$ is said to be $(\varepsilon,\delta)$-differentially private 
if for all set of events $S \subseteq \text{Range}(\mathcal{M})$,
for all neighboring data sets $D,D' \in \mathcal{D}^n$ (where $\mathcal{D}$ is the set of all possible data points ) that differ in only one sample:
\begin{align*}
\Pr[\mathcal{M}(D) \in S ] \leq e^{\varepsilon}  \Pr[\mathcal{M}(D') \in S]  + \delta.
\end{align*}
Informally, this definition requires the probability \emph{any} adversary can observe
a difference between the algorithm operating on dataset $D$ versus a neighboring dataset $D'$ 
is bounded by $e^\varepsilon$, plus a constant additive failure rate of $\delta$.
To provide a meaningful theoretical guarantee, $\varepsilon$ is typically set to
small single digit numbers, and $\delta \ll 1/|D|$.
Other variants of DP use slightly different formulations (e.g., R\'enyi
differential privacy \cite{mironov2017renyi} and concentrated DP~\cite{dwork2016concentrated,bun2016concentrated}); our paper studies predominantly this
$(\varepsilon,\delta)$-definition of DP as it is the most common. Furthermore,  it is often possible to translate a property achieved under one formulation to another, in fact
the DP-SGD analysis uses a R\'enyi bound as an intermediate step to achieving the $(\varepsilon,\delta)$
bounds.

Differential privacy has two useful properties we will use. 
First, the composition of multiple differentially private algorithms is still differentially private
(adding their respective privacy budgets).
Second, differential privacy is immune to post-processing: the output of any DP algorithm can
be arbitrarily post-processed while retaining the same privacy guarantees.


Consider the toy problem of reporting the approximate sum of $D = \{x_1, \dots, x_n\}$ with each $x_i \in [-1, 1]$.
%
A standard way to compute a DP sum is the \emph{Gaussian mechanism}~\cite{dwork2014algorithmic}:
\begin{align}
\label{eqn:gaussian}
\mathcal{M}(D)= \sum_{x_i \in D} x_i + Z \hspace{0.5cm}\text{where } Z\sim \mathcal{N}(0,\sigma^2)
\end{align}
If $\sigma=\sqrt{
2 \log(1.25/\delta)}/\varepsilon$, then it can be shown that $\mathcal{M}$ guarantees $(\epsilon,\delta)$-differential privacy---because
each sample has a bounded range of $1$.
If $x_i$ was unbounded, for any fixed amount of noise $\sigma$, it would
always be possible to let
\begin{align*}
\hat{D} = \left\{ 2\sigma, 0, \dots, 0 \right\}, \hspace{.5cm} 
\tilde{D} = \left\{ -2\sigma, 0, \dots, 0 \right\}
\end{align*}
so with high probability $\mathcal{M}(D) > 0$ if and only if $D = \hat{D}$.

\subsection{Differentially-Private Stochastic Gradient Descent}\label{sec:dp_background}

Stochastic gradient descent (SGD) can be made differentially private through two straightforward modifications.
Proceed initially as in SGD, and sample a
minibatch of examples randomly from the training dataset.
Then, as in Equation~\ref{eqn:sgd}, compute the gradient of the loss on this mini-batch with respect
to the model parameters $\theta$.
However, before directly applying this gradient $\nabla_\theta \mathcal{L}(f_{\theta}, B)$
DP-SGD first makes the gradient differentially private.
Intuitively, we achieve this by 
(1) bounding the contribution of any individual
training example, and then (2) adding a small amount of noise.
Indeed, this can be seen as an application of the Gaussian mechanism to the gradients updates.

To begin, DP-SGD \text{clips} the gradients so that any individual update is bounded in magnitude by $b$,
and then adds Gaussian noise whose scale $\sigma$ is proportional to $b$.
After sampling a minibatch $B=\mathbb{B}(\mathcal{X})$, the new update rule becomes
\begin{equation}
    \label{eqn:dpsgd}
    \theta_{i+1} \gets \theta_{i} - \eta \left({1 \over |B|}\sum_{(x,y) \in B}  \text{clip}_{b}\left(\nabla_{\theta}\ell(f_{\theta_i}, x, y)\right) + Z_i\right)
\end{equation}
where $Z_i \sim \mathcal{N}(0, \sigma^2 \mathbb{I})$
and $\text{clip}_b(v)$ is the function that projects $v$ onto the $\ell_2$ ball of
radius $b$ with
\[\text{clip}_{b}(v) = v \cdot \min\left\{1, \frac{b}{\|v\|_2}\right\}.\]
It is not difficult to see how one would achieve $(\varepsilon,\delta)$-DP
guarantees following Equation~\ref{eqn:gaussian}. To achieve  $(\varepsilon,\delta)$-DP, typically $\sigma$ is in the order of $\Omega(q\frac{\sqrt{T\log(\frac{1}{\delta})}}{\varepsilon})$~\cite{abadi2016deep}. Moreover, Mironov et. al.~\cite{mironov2019r} showed tighter bounds for a given $\sigma$. 
%
On each iteration,
we are computing a differentially-private update of the model parameters given the
gradient update $\nabla_\theta\ell$ through the subsampled Gaussian Mechanism.
Then, because DP is immune to post-processing, we can apply these gradients to
the model.
Finally, through composition, we can obtain a guarantee for the entire model training pipeline.

However, it turns out that naive composition using this trivial analysis
gives values of $\varepsilon \gg 10^4$ for accurate neural networks,
implying a ratio of adversary true positive to false positive bounded by $e^{10^4}$---and hence not meaningful.
Thus, Abadi \emph{et al.} \cite{abadi2016deep} develop the Moments Accountant: an
improvement of Bassily \emph{et al.} \cite{bassily2014private} which introduces a
much more sophisticated tool to analyze DP-SGD that can prove, \emph{for the same
algorithm}, values of $\varepsilon < 10$.\fix{ Since then, many works~\cite{dong2019gaussian,asoodeh2020better,wang2019subsampled,mironov2019r,koskela2020tight,balle2018privacy} improved the analysis of the Moments Accountant analysis to get better theoretical privacy bounds.}

This raises our question: 
is the current analysis the best
analysis tool we could hope for?
Or does there exist a stronger analysis approach that could, again for the same
algorithm---but perhaps with different (stronger) assumptions---reduce $\varepsilon$ by orders of magnitude?

\section{Analysis Approach: Adversary Instantiation}

We answer this question and analyze the tightness of the
DP-SGD analysis under various assumptions by \emph{instantiating the adversary} against the differentially private training algorithm.
This gives us a lower bounds on the privacy leakage that may result.
Our primary goal is to not only investigate the tightness of DP-SGD under any \fix{one particular set of assumptions \cite{jagielski2020auditing}}, but also to understand the relative importance of adversary capabilities assumed to establish this upper bound.
This allows us to appreciate what practical conditions need to be met before an adversary exploits  the full extent of the privacy leakage tolerated by the upper bound. 
This section develops our core algorithm: \emph{adversary instantiation}.
We begin with our motivations to construct this adversary (\S~\ref{sec:motivation})
present the algorithm (\S~\ref{sec:algorithm}) and then 
describe how to use this algorithm to lower bound privacy (\S~\ref{sec:lowerbound}).


\begin{figure*}
    \centering
    \includegraphics[scale=0.8]{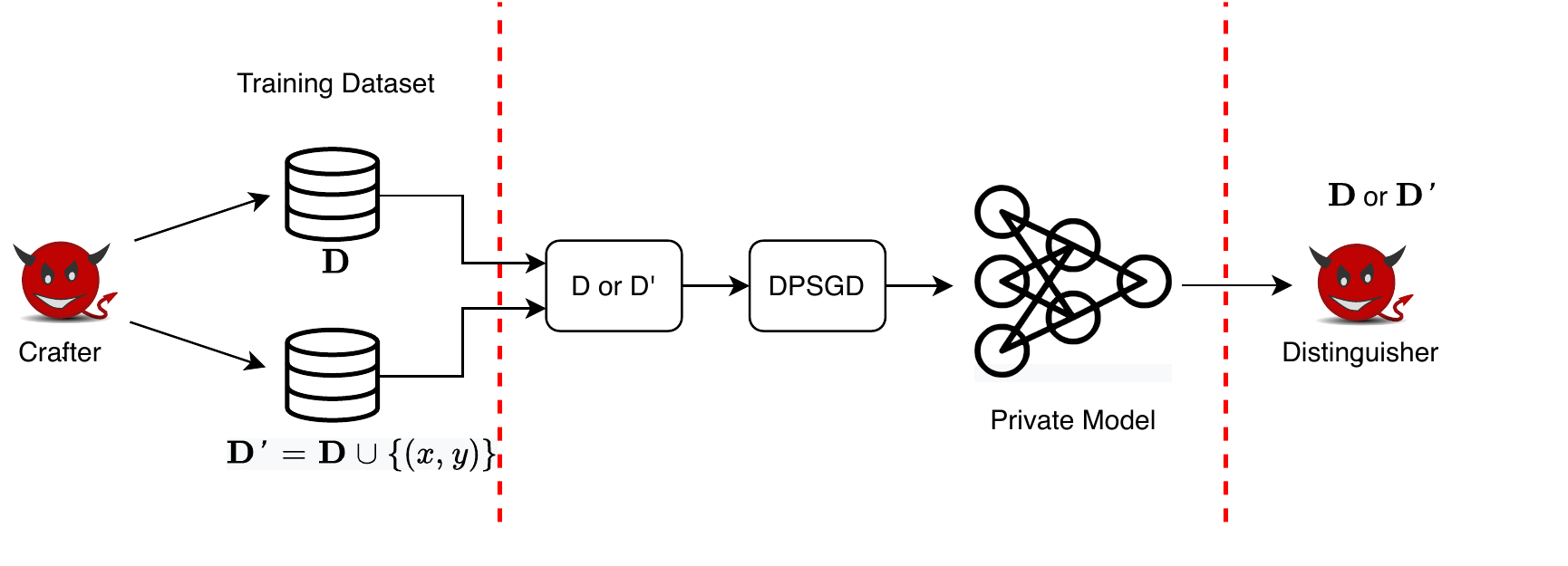}
    \caption{\textbf{Our attack process.} Our first algorithm, the \textbf{crafter} constructs
    two datasets $D$ and $D'$ differing in one example. The model trainer  then (independently of the adversary) trains a model on
    one of these two datasets. Our second algorithm, the \textbf{distinguisher} then guesses
    which dataset was used.}
    
    %
    
    %
    
    \label{fig:overview}
\end{figure*}

\subsection{Motivation: Adversary Capabilities}
\label{sec:motivation}

The analysis of DP-SGD is an over-approximation of actual adversaries
that occur in practice.
This is for two reasons.
First, there are various assumptions that DP-SGD requires that are necessarily imposed because DP captures more powerful adversaries than typically exist.
Second, there are adversary restrictions that we would immediately 
choose to enforce if there was a way to do so, but currently have no way
to make use of in the privacy analysis.
We briefly discuss each of these.

\paragraph{Restrictions imposed by DP}
%
Early definitions of privacy were often dataset-dependent definitions.
For example, $k$-anonymity argues (informally) that a data sample is private if there
are at least $k$ ``similar'' \cite{sweeney2002k}.
Unfortunately, these privacy definitions often have problems when dealing
with high dimensional data \cite{narayanan2006break}---allowing attacks that nevertheless reveal sensitive information.

In order to avoid these difficulties, differential privacy makes
dataset-agnostic guarantees: $(\varepsilon,\delta)$-DP must
hold \emph{for all} datasets $D$, even for pathologically constructed datasets $D$.
It is therefore conceivable that the total privacy afforded
to any user in a ``typical'' dataset might be higher than
if they were placed in the worst-case dataset for that user.

\paragraph{Restrictions imposed by the analysis}
The second class of capabilities are those
allowed by the proof that the DP-SGD update rule (Equation~\ref{eqn:dpsgd})
satisfies differential privacy, but if it were possible to improve
the analysis by preventing these attacks it would be done.
Unfortunately, there is no known way to improve on the analysis by prohibiting
these particular capabilities---even if we believe that it should help.

The canonical example of this is the publication of intermediate models.
The DP-SGD analysis assumes the adversary is given all
intermediate model updates $\{\theta_i\}_{i=1}^N$ used to train the neural
network, and not just the final model.
These assumptions is not made because it is necessary to satisfying
differential privacy, but because the only way we know how to analyze DP-SGD
is through a composition over mini-batch iterations, where the adversary
learned all intermediate updates.

In the special case of training convex models, it turns out there
is a way to \emph{directly} analyze the privacy of the final learned model~\cite{feldman2018privacy}.
This gives a substantially stronger guarantee than can be proven by
summing up the cumulative privacy loss over each iteration of the training
algorithm.
However, for general deep neural networks, there are
no known techniques to achieve this. Hence, we ask the question: does having access to the intermediate model updates allow an adversary to mount an attack that leaks more private information from a deep neural network's training set? 

Similarly, the analysis in differential privacy assumes
the adversary has direct control of the gradient updates
to the model.
Again however, this is often not true in practice:
the inputs to a machine learning training pipeline are
input examples, not gradients.\footnote{For
\emph{federated learning} \cite{mcmahan2017communication} an adversary will have this capability because participants in this   protocol contribute to learning by sharing gradient updates.}
The gradients are derived from the current model and the
worst-case inputs.
The proofs place the
the trust boundary at the gradients
simply because it is simpler to analyze the system this way.
If it was possible to analyze the algorithm at the more
realistic interface of examples, it would be done. Through our work, we are able to show when doing so would not improve the upper bound, in which case it would be unnecessary to attempt an improvement of the analysis that relaxes this assumption.

\subsection{Instantiating the DP Adversary}
\label{sec:algorithm}

Because assumptions made to analyze the privacy of DP-SGD appear conservative, researchers
often conjecture that the privacy afforded by DP-SGD is significantly
stronger than what can be proven.
For example, Carlini \emph{et al.} \cite{carlini2019secret} argue (and provide some evidence) 
that training language 
models with $\varepsilon=10^5$-DP might be safe in practice despite this
offering no theoretical guarantees. In other words, it is assumed that there is a gap between the lower bound an adversary may achieve when attacking the training algorithm and the upper bound established through the analysis. 

Before we expend more effort as a community to improve on the DP-SGD analysis to tighten the upper bound, it is important to investigate how wide it may be. If we could show that the gap is non-existent, there would be no point in trying to tighten the analysis, and instead, we would need to identify additional constraints that could be placed on the adversary  in order to decrease the worst-case privacy leakage.
This motivates our alternate method to measure privacy with a lower bound, through developing an attack.
This lets us directly measure
how much privacy restricting each of these currently-allowed
capabilities would afford, should we be able to analyze the resulting training algorithm strictly.

As the core technical contribution of this paper,
in order to  measure this potential gap left by the privacy analysis of DP-SGD, 
we instantiate the adversary introduced in the formulation of differential privacy.
The objective of the theoretical DP adversary is to distinguish between a model $f$ that was
either trained on dataset ${D}$ or on dataset ${D}'$, 
where the datasets only differ at exactly one instance.
Thus, our instantiated adversary consists of two algorithms:
the first algorithm chooses the datasets ${D}$ and ${D'}$,
after which a model is trained on one of these at random,
and then the second algorithm predicts which dataset was selected and trained on.
Figure~\ref{fig:overview} gives a schematic of our attack
approach containing the three phases of our attack,
described in detail below.

\paragraph{The \alga} The \alga{} adversary is a randomized algorithm taking no inputs and returning two possible sets
of inputs to model training algorithm $\mathcal{T}$. Intuitively, this adversary corresponds to  the differential privacy analysis being data-independent: it should hold for all pairs of datasets which only differ by one point.
We will propose concrete implementations later in Section~\ref{sec:attack1} through \ref{sec:attack6}.
Formally, we denote this adversary by the function $\mathcal{A}$.
In most cases, this means that $\mathcal{A} \to ({D}, (x^*, y^*))$,
with ${D'} = {D} \cup \{(x^*, y^*)\}$.

\paragraph{Model training (not adversarial).}
After the \alga{} has supplied the two datasets,
the training algorithm runs in a black box outside of the control of the adversary.
The model trainer first randomly chooses one of the two datasets supplied by the adversary, 
and then trains the model on this dataset (either for one step or the full training run).
Depending on the exact details of the training algorithm, differing levels of information are revealed to the adversary.
For example, the standard training algorithm in DP-SGD is assumed to reveal all intermediate
models $\{\theta_i\}_{i=1}^N$.
We denote training by the randomized algorithm $t = \mathcal{T}({D})$.

\paragraph{The \algb} The \algb{} adversary is a randomized algorithm $\mathcal{B} \to \{0,1\}$
that predicts $0$ if $\mathcal{T}$ was trained on ${D}$
(as produced by $\mathcal{A}$), and $1$ if ${D}'$.
This corresponds to the differential privacy analysis which estimates how much an adversary can improve its odds of guessing whether the training algorithm learned from $D$ or $D'$.

The \algb{} receives as input two pieces of data: the output of \alga{}, and the
output of the model training process.
Thus, in all cases the \algb{} receives the two datasets produced by \alga{},
however in some setups when the \alga{} is called multiple times,
the \algb{} receives the output of all runs.
Along with this, the \alga{} receives the output of the training process, which
again depends on the experimental setup.
In most cases, this is either the final trained model weights $\theta_N$ or the
sequence of weights $\{\theta_i\}_{i=1}^N$.

\subsection{Lower Bounding $\varepsilon$}
\label{sec:lowerbound}
Performing a single run of the above protocol gives a single bit: either the adversary wins
(by correctly guessing which dataset was used during training) or not (by guessing incorrectly).
In order to be able to provide meaningful analysis, we must repeat the above attack multiple times
and compute statistics on the resulting set of trials.

We follow an analysis approach previously employed to audit DP~\cite{ding2018detecting,jagielski2020auditing}.
We extend the analysis to be able to reason about both pure DP where $\delta=0$ and the more commonly employed $(\varepsilon, \delta)$ variant, unlike prior work which assumes $\delta$ is always negligible~\cite{jagielski2020auditing}.
We define one \emph{instance} of the attack as the \alga{} choosing a pair of datasets,
the model being trained on one of these, and the \algb{} making its guess.
For a fixed training algorithm we would like to run, we define a pair of adversaries.
Then, we run a large number of instances on this problem setup.
Through Monte Carlo methods, we can then compute a lower bound on the $(\varepsilon,\delta)$-DP.

Given the success or failure bit from each instance of the attack, we compute the false positive
and false negative rates.
Defining the positive and negative classes is arbitrary; without loss of generality we define a
a false positive as the adversary guessing ${D}'$ when the model was trained on ${D}$,
and vice versa for a false negative.
Kairouz et. al~\cite{kairouz2015composition} showed if a mechanism $\mathcal{M}$ is $(\epsilon,\delta)$-differentially private then the adversary's false positive (FP) and false negative (FN) rate is bounded by:
\begin{align}
    FP + e^{\varepsilon} FN \leq 1-\delta \nonumber \hspace{3em}
    FN + e^{\varepsilon} FP \leq 1-\delta \nonumber
\end{align}

Therefore, given an appropriate $\delta$, we can determine the empirical $(\varepsilon,\delta)-$differential privacy as
\begin{align}\label{eq:attack_dp}
    \varepsilon_{empirical} = \max( \log \frac{1-\delta- FP}{FN}, \log \frac{1-\delta- FN}{FP} )
\end{align}
Using the Clopper-Pearson method~\cite{clopper1934use}, we can determine confidence bounds on the attack performance.
This lets us determine a lower confidence bound on the empirical $\varepsilon$ as
\begin{align}\label{eq:attack_dp_lower}
    \varepsilon_{empirical}^{lower} = \max( \log \frac{1-\delta- FP^{high}}{FN^{high}}, \log \frac{1-\delta- FN^{high}}{FP^{high}} )
\end{align}
where $\mathit{FP}^{high}$ is the high confidence bound for false positive and $\mathit{FN}^{high}$ is the high confidence bound for false negative.
%

We repeat most of our experiments 1000 times in order to measure the FPR and FNR.
Even if the adversary were to succeed in all 1000 of the trials, the Clopper-Pearson bound would 
imply an epsilon lower bound of 5.60.
Note that as a result of this, it will never be possible for us to establish a lower bound greater than $\varepsilon=5.60$ with 1000 trials and a confidence bound of 
due solely to statistical power.

\begin{table*}[ht]
    \centering
    \begin{tabular}{cc|ccc|ccc|c}
    \toprule
        \textbf{Section} & \textbf{Experiment}  & \textbf{Access} & \textbf{Modification} & \textbf{Dataset} & \textbf{Protocol} & \multicolumn{2}{c}{\textbf{Adversaries}} & \textbf{Results}  \\
        \midrule
        \S~\ref{sec:attack1} & API access & Black-box &   Random Sample  &  Not modified  & \cref{fig:protocol1} & \cref{alg:craft_mi} & \cref{alg:dist_mi} & \cref{fig:mi_sgd} \\
        \S~\ref{sec:attack2} & Static Poison Input & Final Model &   Malicious Sample  &  Not modified  & \cref{fig:protocol1} & \cref{alg:craft_malicious} & \cref{alg:dist_mi} & \cref{fig:static_malinput_sgd} \\
        \S~\ref{sec:attack3} & Intermediate Attack & Intermediate Models &   Malicious Sample  &  Not modified      & \cref{fig:protocol2} & \cref{alg:craft_malicious} & \cref{alg:MI_wb} & \cref{fig:static_malinput_sgd_white} \\
        \S~\ref{sec:attack4} & Adaptive Poison Input & Intermediate Models &   Adaptive Sample  &  Not modified     & \cref{fig:protocol4} & \cref{alg:craft_malicious_dynamic} & \cref{alg:MI_wb} & \cref{fig:adap_sgd}\\
        \S~\ref{sec:attack5} & Gradient Attack  & Intermediate Models &   Malicious Gradient  &  Not modified    & \cref{fig:protocol5} & \cref{alg:gradient} & \cref{alg:gradient_detection} & \cref{fig:malgrad_sgd}\\
        \S~\ref{sec:attack6} & Dataset Attack  & Intermediate Models &   Malicious Gradient  &  Malicious    & \cref{fig:protocol5} & \cref{alg:mal_data} & \cref{alg:gradient_detection} & \cref{fig:grad_attack_fgd-maldata} \\
        \bottomrule

        %
        
    \end{tabular}
    \caption{\textbf{Experiment settings.} We run attacks under six set of adversary capabilities;
    each experiment increases the capabilities of the adversary ranging from the weakest membership
    inference adversary to the strongest adversary with full DP capabilities.}
    \label{tab:diff_settings}
\end{table*}

\section{Experiments}

Having developed the methodology to establish a lower bound on
differential privacy guarantees, we now apply it across
six attack models, where we vary  adversary capabilities around four orthogonal key aspects of the ML pipeline.
\begin{itemize}
    \item \textbf{Access.} What level of access does the adversary have to the output of the training algorithm?
    \item \textbf{Modification.} How does the adversary perform the manipulations of the training dataset?
    \item \textbf{Dataset.} Does the adversary control the entire dataset? Or just one poisoned example?
    Or is the dataset assumed to be a typical dataset (e.g., CIFAR10)?
\end{itemize}

While it would (in principle) 
be possible to evaluate the effect of every possible
combination of the assumptions,
it is computationally intractable.
Thus, we describe six possible useful configurations, where by useful we mean configurations  which correspond to relevant modern deployments of ML (e.g., MLaaS, federated learning, etc.) or key aspects of the privacy analysis.
Table~\ref{tab:diff_settings} summarizes the
adversaries we consider. Note that some of the attacks use similar \alga{} or \algb{}, we did not repeat the identical algorithms. Please refer to Table~\ref{tab:diff_settings} for the \alga{} or \algb{} for each attack.
Below we describe them briefly and relate them to one another before expanding on them in the remainder of this section.
\fix{Note that we do not argue that any particular set of assumptions can provide a specific privacy bounds, but rather that it cannot be more private than our experimental bounds.}

\paragraphb{API access:} 
The data owner will collect the data and train the model itself, the adversary cannot control and modify the training procedure or dataset. The adversary can only have black box access to the trained model. This setting is the most practical attack: it is a running example in the ML security literature given the increased popularity of the ML as a Service (MLaaS) deployment scenario. 
In this most realistic yet limited evaluation, we establish a baseline lower bound through a membership inference attack~\cite{shokri2017membership}.

\paragraphb{Static Poison Attack:}
While ``typical-case'' privacy leakage is an important factor in understanding privacy risk,
it is also worth evaluating the privacy of worst-case outliers.
For example, when training a model on a dataset containing exclusively individuals
of one type (age, gender, race, etc), it is important to understand if inserting a training example 
from an under-represented group would increase the likelihood of private information leaking from the model.
To study this setting, we instantiate our adversary to construct worst-case malicious
inputs, and has access to the final trained model.

\paragraphb{Intermediate Poison Attack:}
As mentioned earlier, the DP-SGD analysis assumes the adversary is given access to all
intermediate model parameters. 
We repeat the above attack, but this time reveal all intermediate models.
This allows us to study the relative importance of this additional capability.

\paragraphb{Adaptive Poison Attack:}
The DP-SGD analysis does not have a concept of a dataset: it operates solely on
gradients computed over minibatches of training data.
Thus, there is no requirement that the one inserted example
remains the same between epochs: at each iteration, we re-instantiate the
worst-case inserted example after each epoch.
This again allows us to evaluate whether the minibatch perspective taken in the analysis affects the lower bound we are able to provide.

\paragraphb{Gradient Attack:}
As federated learning continues to receive more attention, it is also important to estimate how much additional privacy leakage results from training a model in this environment. In federated learning a malicious entity can poison the gradients themselves. As mentioned in Section~\ref{sec:dp_background}, DP-SGD also assumes the adversary  has access to the gradients, therefore we mount an attack where the adversary  modifies the gradient directly.

\paragraphb{Dataset Attack:}
In our final and most powerful attack, we make use of all adversary capabilities.
The DP-SGD analysis yields a guarantee which holds for \emph{all} pairs of datasets with a Hamming distance of 1, even if these datasets are  pathological worst-case datasets.
Thus, in this setting, we allow the adversary to construct such a pathalogical dataset.
Evaluating this setting is what allows us to demonstrate that the DP-SGD analysis is tight. \fix{The tightness of the DP-SGD privay bounds in the worst case can also be inferred from the theoretical analysis~\cite{dong2019gaussian,mironov2019r}, however, the main purpose of this evaluation is to show that our attack can be tight when the adversary is the most powerful. If we could not reach the provable upper bound, then we would have no hope that our results would be tight in the other settings. Moreover, we crafted a specific attack to achieve the highest possible privacy leakage.}

\subsection{API Access Adversary}
\label{sec:attack1}

As a first example of how our adaptive analysis approach proceeds, we consider the
baseline adversary who mounts the well-researched membership inference attack \cite{shokri2017membership} in a setting where they have access to an API. As described above, this corresponds to the most practical setting---a black-box attack where the attacker is only given access to an API revealing the model's output (its confidence so has to be able to compute the loss) on inputs chosen by the attacker.
The schematic for this analysis is given in \cref{fig:protocol1}.
%
\newcommand\sample{\stackrel{\mathclap{\scriptsize\mbox{\$}}}{\gets}}

\begin{protocol}
\centering
\begin{tabular}{lcr}
\multicolumn{3}{c}{\textbf{Membership Inference Adversary Game}} \\
\textbf{Adversary} & \hspace{3em} & \textbf{Model Trainer} \\
\midrule
$(D,D') \sample \mathcal{A} $ & & $b \sample \{0, 1\}$ \\
\multicolumn{3}{c}{\sendmessageright{top={$B = (D,D)$}}} \\
& & $f_\theta \gets \mathcal{T}(B_b)$ \\
\multicolumn{3}{c}{\sendmessageleft{top={$f_\theta$}}} \\
$s \gets \mathcal{B}(f_\theta, D, x^*, y^*) $ & & \\
\multicolumn{3}{c}{\sendmessageright{top={$s \in \{0,1\}$}}} \\
& & Check if $s=b$ \\
\end{tabular}
\caption{\textbf{Adversary game for membership inference attack.} 
    \textbf{Round 1. }The adversary chooses two datasets.
    \textbf{Round 2. }The model trainer randomly trains a model on one of these, and returns the
    model. 
    \textbf{Round 3. }The adversary predicts which dataset was used for training.}
    \label{fig:protocol1}
\end{protocol}

\paragraph{\alga} (\cref{alg:craft_mi})
Given any standard machine learning dataset ${D}$, we can remove a random
example from the dataset to construct a new dataset ${D}'$
that differs from ${D}$ in exactly one record.
Formally, let $(x,y) \sample \mathcal{D}$ be a sample from the underlying data distribution at random.
We let $\mathcal{A} = \{D , D\cup \{(x,y)\} \}$,
We formalize this in~\cref{alg:craft_mi}. 

\begin{crafter}
\small
 \caption{Membership Inference Adversary}
 \begin{algorithmic}[1]
  \REQUIRE existing training dataset ${D}$, underlying data distribution $\mathcal{D}$
  \STATE $(x,y) \sample \mathcal{D}$
  \STATE ${D}' \gets {D} \cup \{(x,y)\}$
 \RETURN ${D}, {D}'$
 \end{algorithmic}
 \label{alg:craft_mi}
 \end{crafter}

\paragraph{Model training} Recall from Section~\ref{sec:implementation} that
the trainer gets two datasets: $D$ and $D'$.
Then the trainer selects one of these datasets randomly and trains a model on the selected dataset using pre-defined hyperparameters. After training completes, the trainer outputs the trained model $f_{\theta}$.

\paragraph{\algb} (\cref{alg:dist_mi})
After the model $f_\theta$ has been trained, the \algb{} now guesses which dataset was used by computing the loss of the trained 
model on the one differing example
$\ell(f_\theta, x, y)$.
It guesses that the model was trained on ${D}'$ if the loss is sufficiently
small, and guess that it was trained on ${D}$ otherwise. In the rest of the paper, we refer to the differing example $(x,y)$ as the query input.
%
%
Details are given in \cref{alg:dist_mi}.

\begin{distinguisher}
\small
 \caption{Membership Inference Adversary}
 \begin{algorithmic}[1]
  \REQUIRE model $f_\theta$, query input $(x,y)$, threshold $\tau$
  \STATE $L \gets \ell(f_\theta,x,y)$
  \IF{ $L \leq \tau$}
  \RETURN $D'$
  \ENDIF
  \RETURN $D$
 \end{algorithmic}
 \label{alg:dist_mi}
 \end{distinguisher}

Previous works~\cite{shokri2015privacy,nasr2019comprehensive,melis2019exploiting,choo2020label} showed that if an instance is part of the training dataset of the target model, its loss is most likely less than the case when it is not part of the training dataset.  
That is, in this attack, we provide our instantiated adversary with an extremely
limited power. In particular, the attacker has much less capabilities than what is assumed to analyze the differential privacy guarantees provided by DP-SGD. 
\cref{fig:protocol1} summarizes the attack.

\paragraph{Results}
Typically, membership inference is studied as an attack in order to
learn if a given
user is a member of the training dataset or not.
However, we do not study it as an attack, but as a way to provide a lower bound on the privacy leakage that arises from training with DP-SGD.

We experiment with a neural network
using three  datasets commonly used in privacy research: MNIST~\cite{lecun1990handwritten}, CIFAR-10~\cite{krizhevsky2009learning}, and Purchase.
The first two of these are standard image classification datasets, and Purchase is the shopping records of several thousand online customers, extracted
during Kaggle’s ``acquire valued shopper'' challenge \cite{kaggle}.
Full details of these datasets is given in Appendix~\ref{sec:appendix_dataset}

In each trial, we follow  \cref{fig:protocol1} for each of the
three datasets.
We train a differentially private model using DP-SGD setting $\varepsilon$
to typical values used in the machine learning literature: 1, 2, 4, and 10.
As mentioned before,
the more trials we perform the better we are able to establish a lower bound of $\varepsilon$.
Due to computational constraints, we are limited to performing the attack $1,000$ times.
These experiments took 3,000 GPU hours, parallelized over 24 GPUs.

When we perform this attack on the CIFAR-10 dataset and train a model with
$\varepsilon=4$ differential privacy, the attack true positive rate is $0.017$ and false positive is $0.002$.
By using the Clopper-Pearson method, we can probabilistically lower-bound this attack performance;
for example, when we have performed $1000$ trials, there is a $95\%$ probability that the
attack false positive rate is lower than $0.01$.
Using Equations~\ref{eq:attack_dp} and~\ref{eq:attack_dp_lower},  we can convert \emph{empirical} lower bound of $(\varepsilon,\delta)$-DP
with $(0.31,10^{-5})$.
This value is \emph{substantially} lower than the \emph{provably correct}
upper bound of $(4,10^{-5})$.

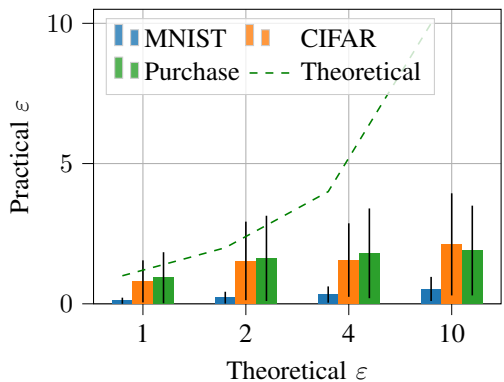
\begin{figure}[t]
    \centering
\begin{tikzpicture}

\definecolor{color0}{rgb}{0.12156862745098,0.466666666666667,0.705882352941177}
\definecolor{color1}{rgb}{1,0.498039215686275,0.0549019607843137}
\definecolor{color2}{rgb}{0.172549019607843,0.627450980392157,0.172549019607843}

\begin{axis}[
height=\figureheight,
width=\figurewidth,
legend cell align={left},
legend columns=2,
legend style={fill opacity=0.8, draw opacity=1, text opacity=1, at={(0.03,0.97)}, anchor=north west, draw=white!80!black},
tick align=outside,
tick pos=left,
x grid style={white!69.0196078431373!black},
xlabel={Theoretical \(\displaystyle  \varepsilon\)},
xmajorgrids,
xmin=-0.28, xmax=3.68,
xtick style={color=black},
xtick={0.2,1.2,2.2,3.2},
xticklabels={1,2,4,10},
y grid style={white!69.0196078431373!black},
ylabel={Practical \(\displaystyle  \varepsilon\)},
ymajorgrids,
ymin=0, ymax=10.5,
ytick style={color=black}
]
\draw[draw=none,fill=color0] (axis cs:-0.1,0) rectangle (axis cs:0.1,0.11);
\addlegendimage{ybar,ybar legend,draw=none,fill=color0};
\addlegendentry{MNIST}

\draw[draw=none,fill=color0] (axis cs:0.9,0) rectangle (axis cs:1.1,0.22);
\draw[draw=none,fill=color0] (axis cs:1.9,0) rectangle (axis cs:2.1,0.33);
\draw[draw=none,fill=color0] (axis cs:2.9,0) rectangle (axis cs:3.1,0.53);
\draw[draw=none,fill=color1] (axis cs:0.1,0) rectangle (axis cs:0.3,0.8);
\addlegendimage{ybar,ybar legend,draw=none,fill=color1};
\addlegendentry{CIFAR}

\draw[draw=none,fill=color1] (axis cs:1.1,0) rectangle (axis cs:1.3,1.53);
\draw[draw=none,fill=color1] (axis cs:2.1,0) rectangle (axis cs:2.3,1.56);
\draw[draw=none,fill=color1] (axis cs:3.1,0) rectangle (axis cs:3.3,2.12);
\draw[draw=none,fill=color2] (axis cs:0.3,0) rectangle (axis cs:0.5,0.93);
\addlegendimage{ybar,ybar legend,draw=none,fill=color2};
\addlegendentry{Purchase}

\draw[draw=none,fill=color2] (axis cs:1.3,0) rectangle (axis cs:1.5,1.62);
\draw[draw=none,fill=color2] (axis cs:2.3,0) rectangle (axis cs:2.5,1.8);
\draw[draw=none,fill=color2] (axis cs:3.3,0) rectangle (axis cs:3.5,1.9);
\path [draw=black, semithick]
(axis cs:0,0.001)
--(axis cs:0,0.219);

\path [draw=black, semithick]
(axis cs:1,0.01)
--(axis cs:1,0.43);

\path [draw=black, semithick]
(axis cs:2,0.04)
--(axis cs:2,0.62);

\path [draw=black, semithick]
(axis cs:3,0.1)
--(axis cs:3,0.96);

\path [draw=black, semithick]
(axis cs:0.2,0.05)
--(axis cs:0.2,1.55);

\path [draw=black, semithick]
(axis cs:1.2,0.13)
--(axis cs:1.2,2.93);

\path [draw=black, semithick]
(axis cs:2.2,0.25)
--(axis cs:2.2,2.87);

\path [draw=black, semithick]
(axis cs:3.2,0.3)
--(axis cs:3.2,3.94);

\path [draw=black, semithick]
(axis cs:0.4,0.02)
--(axis cs:0.4,1.84);

\path [draw=black, semithick]
(axis cs:1.4,0.1)
--(axis cs:1.4,3.14);

\path [draw=black, semithick]
(axis cs:2.4,0.2)
--(axis cs:2.4,3.4);

\path [draw=black, semithick]
(axis cs:3.4,0.3)
--(axis cs:3.4,3.5);

\addplot [semithick, green!50!black, dashed]
table {%
0 1
1 2
2 4
3 10
};
\addlegendentry{Theoretical}
\end{axis}

\end{tikzpicture}
    \caption{\textbf{Membership inference attack:} the adversary only adds one sample from the underlying data distribution. }
    \label{fig:mi_sgd}
\end{figure}

Figure~\ref{fig:mi_sgd} shows the empirical epsilon for the other datasets and values of differential privacy. 
For the MNIST dataset, the adversary's advantage is not significantly better than random chance; we hypothesize that this is because MNIST images are all highly similar, and so inserting or removing any one training example from the dataset does not change the model by much.
CIFAR-10 and Purchase, however, are much more diverse tasks and the adversary can distinguish between two dataset ${D}$ and ${D}'$ much more easily. In general, the API access attack does not utilize any of the assumptions assumed to be available to the adversary in the DP-SGD analysis and as a consequence the average input from the underlying data distribution does not leak as much  private information as suggested by the theoretical upper bound. However, as the tasks get more complex information leakage increases.

There are two possible interpretations of this result, which will be the main focus of this paper:
\begin{enumerate}
    \item \textbf{Interpretation 1.} The differentially private upper bound is overly pessimistic.
    The privacy offered by DP-SGD is much stronger than the bound which
    can be proven.
    \item \textbf{Inteerpretation 2.} This attack is weak. A stronger attack could have succeeded more often,
    and as such the bounds offered by DP-SGD might be accurate.
\end{enumerate}

Prior work has often observed this phenomenon, and for example trained
models with $\varepsilon=10^5$-DP \cite{carlini2019secret}
despite this offering almost no theoretical privacy, because
empirically the privacy appeared to be strong.
However, as we have already revealed in the introduction, 
it turns out that Interpretation 2 is correct: 
while, for this weak attack, the bounds are loose,
this does not imply DP-SGD itself is loose when an adversary utilizes all capabilities.

\subsection{Static Input Poisoning Adversary}
\label{sec:attack2}

The previous attack assumes a weak adversary to establish the first baseline lower bound in a realistic attack setting.
Nothing constructed by $\mathcal{A}$ is adversarial \emph{per se}, but rather selected at random from a pre-existing dataset.
This subsection begins to strengthen the adversary.

Differential privacy bounds the leakage when two datasets ${D}$ and ${D}'$  differ in \emph{any instance}. Thus, by following prior work \cite{jagielski2020auditing}, we create an adversary that crafts a poisoned malicious input $x$ such that when the model trains on this input, its output will be different from the case where the instance is not included---thus making membership inference easier.


\paragraph{The \alga{}} (\cref{alg:craft_malicious})
Inspired by Jagielski \emph{et al.} \cite{jagielski2020auditing}, we construct an input designed to \emph{poison} the ML model. 
Given access to samples from the underlying data distribution,
the adversary trains a set of shadow models (e.g., as done in \cite{shokri2017membership}). The adversary trains shadow models using the same hyperparameters as the model trainer will use, which allows adversary to approximate how model trainer's model will behave.
Next, the adversary generates an input with an adversarial example algorithm~\cite{moosavi2016deepfool}.
If a model trains on that input, the learned model be different from a model which is not trained on that input. \cref{alg:craft_malicious} summarizes the algorithm.
%

\begin{crafter}
\small
 \caption{Static Adversary Crafting}
 \begin{algorithmic}[1]
    \REQUIRE train dataset ${D}_{shadow}$, adversarial train steps $T$, input learning rate $s$, model learning rate $lr$, training dataset ${D}$
    \STATE $f^{shadow}_{1},\cdots,f^{shadow}_{n} \gets \mathcal{T}_{dpsgd}({D}_{shadow})$
    \STATE $x,y \sample $ random sample from ${D}_{shadow}$
    \FOR{ $T$ times}
        \STATE $l_{malicious} \gets \frac{1}{n} \sum_{i=1}^n \ell(f^{shadow}_{i},x,y)$
        \STATE $x \gets x + s \nabla_{x}l_{malicious} $
    \ENDFOR
 \STATE ${D}' = {D}\cup \{(x,y)\} $
 \RETURN ${D}, {D}'$
 \end{algorithmic}
 \label{alg:craft_malicious}
 \end{crafter}

\paragraph{Model training} Identical to prior.

\paragraph{The \algb{}} (\cref{alg:dist_mi}) Identical to prior.

\begin{figure}[t]
    \centering
\begin{tikzpicture}

\definecolor{color0}{rgb}{0.12156862745098,0.466666666666667,0.705882352941177}
\definecolor{color1}{rgb}{1,0.498039215686275,0.0549019607843137}
\definecolor{color2}{rgb}{0.172549019607843,0.627450980392157,0.172549019607843}

\begin{axis}[
height=\figureheight,
width=\figurewidth,
legend cell align={left},
legend columns=2,
legend style={fill opacity=0.8, draw opacity=1, text opacity=1, at={(0.03,0.97)}, anchor=north west, draw=white!80!black},
tick align=outside,
tick pos=left,
x grid style={white!69.0196078431373!black},
xlabel={Theoretical \(\displaystyle  \varepsilon\)},
xmajorgrids,
xmin=-0.28, xmax=3.68,
xtick style={color=black},
xtick={0.2,1.2,2.2,3.2},
xticklabels={1,2,4,10},
y grid style={white!69.0196078431373!black},
ylabel={Practical \(\displaystyle  \varepsilon\)},
ymajorgrids,
ymin=0, ymax=10.5,
ytick style={color=black}
]
\draw[draw=none,fill=color0] (axis cs:-0.1,0) rectangle (axis cs:0.1,0.81);
\addlegendimage{ybar,ybar legend,draw=none,fill=color0};
\addlegendentry{MNIST}

\draw[draw=none,fill=color0] (axis cs:0.9,0) rectangle (axis cs:1.1,1.22);
\draw[draw=none,fill=color0] (axis cs:1.9,0) rectangle (axis cs:2.1,1.76);
\draw[draw=none,fill=color0] (axis cs:2.9,0) rectangle (axis cs:3.1,2.51);
\draw[draw=none,fill=color1] (axis cs:0.1,0) rectangle (axis cs:0.3,0.8);
\addlegendimage{ybar,ybar legend,draw=none,fill=color1};
\addlegendentry{CIFAR}

\draw[draw=none,fill=color1] (axis cs:1.1,0) rectangle (axis cs:1.3,1.51);
\draw[draw=none,fill=color1] (axis cs:2.1,0) rectangle (axis cs:2.3,1.56);
\draw[draw=none,fill=color1] (axis cs:3.1,0) rectangle (axis cs:3.3,2.13);
\draw[draw=none,fill=color2] (axis cs:0.3,0) rectangle (axis cs:0.5,0.92);
\addlegendimage{ybar,ybar legend,draw=none,fill=color2};
\addlegendentry{Purchase}

\draw[draw=none,fill=color2] (axis cs:1.3,0) rectangle (axis cs:1.5,1.59);
\draw[draw=none,fill=color2] (axis cs:2.3,0) rectangle (axis cs:2.5,1.94);
\draw[draw=none,fill=color2] (axis cs:3.3,0) rectangle (axis cs:3.5,2.19);
\path [draw=black, semithick]
(axis cs:0,0.01)
--(axis cs:0,1.61);

\path [draw=black, semithick]
(axis cs:1,0.221)
--(axis cs:1,2.219);

\path [draw=black, semithick]
(axis cs:2,0.345)
--(axis cs:2,3.175);

\path [draw=black, semithick]
(axis cs:3,0.43)
--(axis cs:3,4.59);

\path [draw=black, semithick]
(axis cs:0.2,0.05)
--(axis cs:0.2,1.55);

\path [draw=black, semithick]
(axis cs:1.2,0.2)
--(axis cs:1.2,2.82);

\path [draw=black, semithick]
(axis cs:2.2,0.31)
--(axis cs:2.2,2.81);

\path [draw=black, semithick]
(axis cs:3.2,0.41)
--(axis cs:3.2,3.85);

\path [draw=black, semithick]
(axis cs:0.4,0.12)
--(axis cs:0.4,1.72);

\path [draw=black, semithick]
(axis cs:1.4,0.28)
--(axis cs:1.4,2.9);

\path [draw=black, semithick]
(axis cs:2.4,0.4)
--(axis cs:2.4,3.48);

\path [draw=black, semithick]
(axis cs:3.4,0.46)
--(axis cs:3.4,3.92);

\addplot [semithick, green!50!black, dashed]
table {%
0 1
1 2
2 4
3 10
};
\addlegendentry{Theoretical}
\end{axis}

\end{tikzpicture}
    \caption{\textbf{Malicious input attack}: the adversary has blackbox access. The maliciously crafted input leaks more information than a random sample from the data distribution.}
    \label{fig:static_malinput_sgd}
\end{figure}
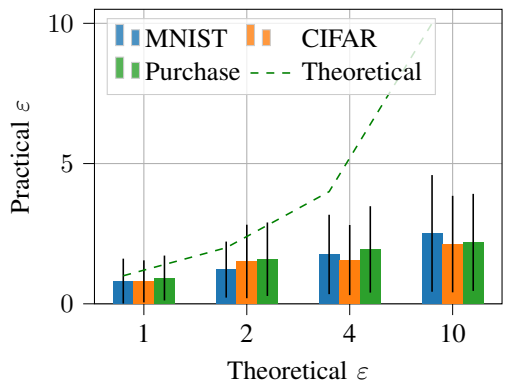

\paragraph{Results} As visualized in
Figure~\ref{fig:static_malinput_sgd}, this adversary is able to leverage its poisoning capability to leak more private information and achieve a  higher empirical lower bound compared to the previous attack. This is consistent  across all datasets.
This is explained by the fact that the poisoned input is a worst-case input for membership inference whereas previously the attack was conducted on average-case inputs drawn from the training distribution. 
The results suggest that DP-SGD bounds are bounded below by a factor of $10\times$.




  

\subsection{Intermediate Poison Attack}
\label{sec:attack3}
The DP-SGD privacy analysis assumes the existence of an adversary who has complete access to the model training pipeline, this includes intermediate gradients computed to update the model parameter values  throughout training.
Instead of   the final model only, we now assume the \algb{} is given access to these intermediate model parameters values, as shown in Protocol~\ref{fig:protocol2}.
(In the case of convex models, it is known that releasing all intermediate
models gives the adversary no more power than just the final model,
however there is no comparable theory for the case of deep neural networks.)

\begin{protocol}
\centering
\begin{tabular}{lcr}
\multicolumn{3}{c}{\textbf{Intermediate Model Adversary Game}} \\
\textbf{Adversary} & \hspace{3em} & \textbf{Model Trainer} \\
\midrule

$(D,D') \sample \mathcal{A} $ & & \\
\multicolumn{3}{c}{\sendmessageright{top={$B = (D,D)$}}} \\
& & $b \sample \{0, 1\}$ \\
& & $\{f_{\theta_i}\}_{i=1}^N \gets \mathcal{T}(B_b)$ \\
\multicolumn{3}{c}{\sendmessageleft{top={$\{f_{\theta_1},f_{\theta_2},\dots,f_{\theta_N}\}$}}} \\
$s \gets \mathcal{B}(\{f_{\theta_i}\}, D, D') $ & & \\
\multicolumn{3}{c}{\sendmessageright{top={$s \in \{0,1\}$}}} \\
& & Check if $s=b$ \\
\end{tabular}
\caption{\textbf{Adversary game for intermediate model adversary.} 
    \textbf{Round 1. }The adversary chooses two datasets.
    \textbf{Round 2. }The model trainer randomly trains a model on one of these, and returns 
    \emph{the full sequence of model updates}.
    \textbf{Round 3. }The adversary predicts which dataset was used for training.}
    \label{fig:protocol2}
\end{protocol}

\paragraph{\alga} (\cref{alg:craft_malicious}) Identical to prior.

\paragraph{Model training}
The model trainer gets the two datasets ${D}$ and ${D}'$ from the \alga. As before we train on one of these two datasets. However, this time, the trainer reveals all intermediate models from the stochastic gradient descent process $\{f_{\theta_i}\}_{i=1}^N$ to the \algb.

\paragraph{\algb} (\cref{alg:MI_wb})
Given the sequence of trained model parameters, we now construct our second adversary to guess
which dataset was selected by the trainer.
We modify Crafter~\ref{alg:dist_mi} to leverage access to the intermediate  outputs of
training. Instead of only looking at  the final model's loss on the
poisoned instance, our adversary also analyzes intermediate losses. We compute either the average or maximum loss for the poisoned examples over all of the intermediate steps and guess the model was trained on $D$
if the loss is smaller than a threshold. 
We took the best attacker between the maximum and average variants. In our experiments, for smaller epsilons ($\varepsilon \leq 2$) the maximum  worked better. For epsilons larger than $2$, it was instead the  average.
Distinguisher~\ref{alg:MI_wb} outlines the resulting attack.

\begin{distinguisher}
\small
 \caption{White-box Membership attack}
 \begin{algorithmic}[1]
    \REQUIRE all intermediate steps to step $T$ $f_1,\cdots,f_T$, query input $(x,y)$, threshold $\tau$, attack method (max or mean)
    \IF{attack method is max}
    \STATE $L \gets \max_{i=1}^T \ell(f_{\theta_i}, x, y)$
    \ENDIF
    \IF{attack method is mean}
    \STATE $L \gets \frac{1}{T} \sum_{i=1}^T \ell(f_{\theta_i}, x, y)$
    \ENDIF
    \IF{ $L < \tau$}
  \RETURN $D'$
  \ENDIF
  \RETURN $D$
 \end{algorithmic}
 \label{alg:MI_wb}
 \end{distinguisher}

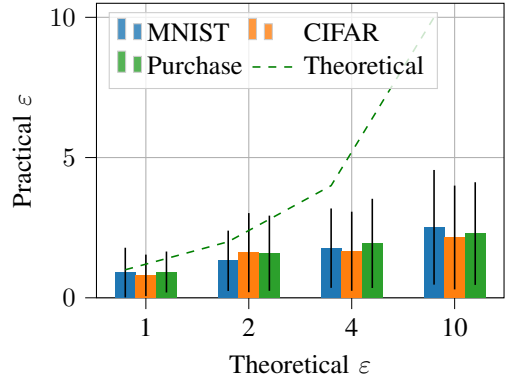
\begin{figure}[t]
    \centering
\begin{tikzpicture}

\definecolor{color0}{rgb}{0.12156862745098,0.466666666666667,0.705882352941177}
\definecolor{color1}{rgb}{1,0.498039215686275,0.0549019607843137}
\definecolor{color2}{rgb}{0.172549019607843,0.627450980392157,0.172549019607843}

\begin{axis}[
height=\figureheight,
width=\figurewidth,
legend cell align={left},
legend columns=2,
legend style={fill opacity=0.8, draw opacity=1, text opacity=1, at={(0.03,0.97)}, anchor=north west, draw=white!80!black},
tick align=outside,
tick pos=left,
x grid style={white!69.0196078431373!black},
xlabel={Theoretical \(\displaystyle  \varepsilon\)},
xmajorgrids,
xmin=-0.28, xmax=3.68,
xtick style={color=black},
xtick={0.2,1.2,2.2,3.2},
xticklabels={1,2,4,10},
y grid style={white!69.0196078431373!black},
ylabel={Practical \(\displaystyle  \varepsilon\)},
ymajorgrids,
ymin=0, ymax=10.5,
ytick style={color=black}
]
\draw[draw=none,fill=color0] (axis cs:-0.1,0) rectangle (axis cs:0.1,0.9);
\addlegendimage{ybar,ybar legend,draw=none,fill=color0};
\addlegendentry{MNIST}

\draw[draw=none,fill=color0] (axis cs:0.9,0) rectangle (axis cs:1.1,1.32);
\draw[draw=none,fill=color0] (axis cs:1.9,0) rectangle (axis cs:2.1,1.77);
\draw[draw=none,fill=color0] (axis cs:2.9,0) rectangle (axis cs:3.1,2.514);
\draw[draw=none,fill=color1] (axis cs:0.1,0) rectangle (axis cs:0.3,0.8);
\addlegendimage{ybar,ybar legend,draw=none,fill=color1};
\addlegendentry{CIFAR}

\draw[draw=none,fill=color1] (axis cs:1.1,0) rectangle (axis cs:1.3,1.61);
\draw[draw=none,fill=color1] (axis cs:2.1,0) rectangle (axis cs:2.3,1.66);
\draw[draw=none,fill=color1] (axis cs:3.1,0) rectangle (axis cs:3.3,2.15);
\draw[draw=none,fill=color2] (axis cs:0.3,0) rectangle (axis cs:0.5,0.92);
\addlegendimage{ybar,ybar legend,draw=none,fill=color2};
\addlegendentry{Purchase}

\draw[draw=none,fill=color2] (axis cs:1.3,0) rectangle (axis cs:1.5,1.59);
\draw[draw=none,fill=color2] (axis cs:2.3,0) rectangle (axis cs:2.5,1.94);
\draw[draw=none,fill=color2] (axis cs:3.3,0) rectangle (axis cs:3.5,2.29);
\path [draw=black, semithick]
(axis cs:0,0.015)
--(axis cs:0,1.785);

\path [draw=black, semithick]
(axis cs:1,0.2451)
--(axis cs:1,2.3949);

\path [draw=black, semithick]
(axis cs:2,0.355)
--(axis cs:2,3.185);

\path [draw=black, semithick]
(axis cs:3,0.47)
--(axis cs:3,4.558);

\path [draw=black, semithick]
(axis cs:0.2,0.0600000000000001)
--(axis cs:0.2,1.54);

\path [draw=black, semithick]
(axis cs:1.2,0.2)
--(axis cs:1.2,3.02);

\path [draw=black, semithick]
(axis cs:2.2,0.25)
--(axis cs:2.2,3.07);

\path [draw=black, semithick]
(axis cs:3.2,0.3)
--(axis cs:3.2,4);

\path [draw=black, semithick]
(axis cs:0.4,0.19)
--(axis cs:0.4,1.65);

\path [draw=black, semithick]
(axis cs:1.4,0.25)
--(axis cs:1.4,2.93);

\path [draw=black, semithick]
(axis cs:2.4,0.35)
--(axis cs:2.4,3.53);

\path [draw=black, semithick]
(axis cs:3.4,0.46)
--(axis cs:3.4,4.12);

\addplot [semithick, green!50!black, dashed]
table {%
0 1
1 2
2 4
3 10
};
\addlegendentry{Theoretical}
\end{axis}

\end{tikzpicture}
    \caption{\textbf{Malicious input attack:} the adversary has white-box access to the training dataset. The results are slightly better compared to the malicious input with blackbox access.}
    \label{fig:static_malinput_sgd_white}
\end{figure}

\paragraph{Results}
Our results in Figure~\ref{fig:static_malinput_sgd_white} show that this adversary only slightly outperforms our previous adversary with access to the final model parameters only. 
This suggests that access to the final model output by the training algorithm leaks almost as much information as the gradients applied during training.
This matches what the theory for convex models suggests, 
even though deep neural networks are non-convex.

\subsection{Adaptive Poisoning Attack}
\label{sec:attack4}

Recall that the DP-SGD analysis treats each iteration of SGD independently, and uses (advanced) composition methods to compute the final privacy bounds. 
As a result, there is no \emph{requirement} that the dataset processed at iteration $i$ need be the same as the dataset used at another iteration $j$.

We design new adversaries to take advantage of this additional capability.
As in the previous attack, our goal is to design two datasets with the property that training on them yields different models.
Again, we will use a query input to distinguish between the models, but for the first time we will not directly place the query input in the training data.
Instead, we will insert a series of (different) poison inputs, each of which is designed to make the models behave differently on the query input.
We describe this process in detail below.
%

\begin{protocol}
\centering
\begin{tabular}{lcrr}
\multicolumn{3}{c}{\textbf{Adaptive Input Poisoning Adversary Game}} \\
\textbf{Adversary}\hspace{2em} & \hspace{3em} & \textbf{Model Trainer} \\
\cmidrule{1-3}
$(x,y) \sample \mathcal{A}_{query}$ & & $b \sample \{0, 1\}$ \\
\multicolumn{3}{c}{\sendmessageleft{top={$f_{\theta_{0}}$}}} \\
\multicolumn{2}{l}{$({D}, D') \sample \mathcal{A}(f_{\theta_i}, x, y)$} & & \rdelim\}{3}{*}[\rotatebox{90}{\hspace{-4em}Repeat $N$ Iterations}] \\
\multicolumn{3}{c}{\sendmessageright{top={$B^i = \{{D}, {D'} \}$}}} \\
& \multicolumn{2}{r}{$\theta_{i+1} \gets \mathcal{S}(\mathbb{B}(B_b^i, x, y), \theta_i)$ }\\
\multicolumn{3}{c}{\sendmessageleft{top={$f_{\theta_{i+1}}$}}} \\
& \vdots & \\

\multicolumn{2}{l}{$s \gets \mathcal{B}(\{f_{\theta_i}\}, {D}, D')$} & & \\
\multicolumn{3}{c}{\sendmessageright{top={$s \in \{0,1\}$}}} \\
& & Check if $s=b$ \\
\end{tabular}
\caption{\textbf{Adversary game for adaptive input poisoning attack.} 
\textbf{Round 0.} The adversary generates a \emph{query input}.
    \textbf{Round 2i. }The \emph{adversary} chooses datasets $D$, $D'$ given the current weights $\theta_i$.
    \textbf{Round 2i+1. }The \emph{model trainer} trains for one minibatch on one of these datasets.
    \textbf{Round N. }The adversary predicts which dataset was used for training.}
    \label{fig:protocol4}
\end{protocol}

\paragraph{\alga} (\cref{alg:craft_malicious_dynamic})
We first generate a \emph{query input} $(x_{q},y_{q})$ by calling \cref{alg:craft_malicious} from the prior section: this is the input that the \algb{} will use to make its guess.
Then, on each iteration of gradient descent, we generate a fresh example $(x,y)$ such that if the model $f$  trains on ${D}\backslash \{(x,y)\}$, we expect that the loss of the query input $x_{q}$ will be significantly larger than training on the full ${D}$, i.e. 
$\ell(f^{{D}\backslash \{(x,y)\}}, x_{q},y_{q}) \ll \ell(f^{{D}}, x_{q},y_{q}).$

To generate the malicious input $x$, we perform double backpropagation.
We compute the gradient of the query input's loss, given a model trained on $x$.
Then, we temporarily apply this gradient update, and then \emph{again} take a gradient this time updating $x$ so that it will minimize the loss of the query input. This process is described in~\cref{alg:craft_malicious_dynamic}.

\begin{crafter}[H]
\small
 \caption{Dynamic Malicious Generation}
 \begin{algorithmic}[1]
    \REQUIRE current model $f_{\theta_T}$, train steps $T$, input learning rate $s$, model learning rate $lr$
    \STATE $(D_a, D'_a) \gets $ \cref{alg:craft_malicious}
    \STATE $x_{query}, y_{query} \gets D'_a \backslash D_a$
    \STATE $x,y \gets $  $\vec{0}$
    \FOR{ $T$ times}
        \STATE $l_{model} \gets \ell(f,x,,y)$
        \STATE $f'(x) \gets f(x) - lr \times \nabla_{f_{\theta_T}} l_{model}  $
        \STATE $l_{malicious} \gets \ell(f', x_{query},y_{query})$
        \STATE $x \gets x + s \nabla_{x}l_{malicious}$
    \ENDFOR
    \STATE ${D}' = {D}\cup \{(x,y)\}$
 \RETURN ${D},{D}'$
 \end{algorithmic}
 \label{alg:craft_malicious_dynamic}
 \end{crafter}

\paragraph{Model training} As usual, the model trainer randomly chooses to train on the benign dataset $D$ or the malicious dataset $D'$. If $D$ is selected,   training proceeds normally. Otherwise, before each iteration of training, the trainer runs ~\cref{alg:craft_malicious_dynamic} first to get an updated malicious $D'$ and selects a mini-batch from the given dataset.  Recall that here, we are interested in the attack for the purpose of establishing a lower bound so while this attack assumes strong control from the adversary on the dataset, it allows us to more accurately bound any potential privacy leakage.

\paragraph{\algb} (\cref{alg:MI_wb})
This adversary is unchanged from the prior section, as described in \cref{alg:MI_wb}.
However, as indicated above, we use the query input $(x_{query},y_{query})$ instead of the poison examples that are inserted into the training data.

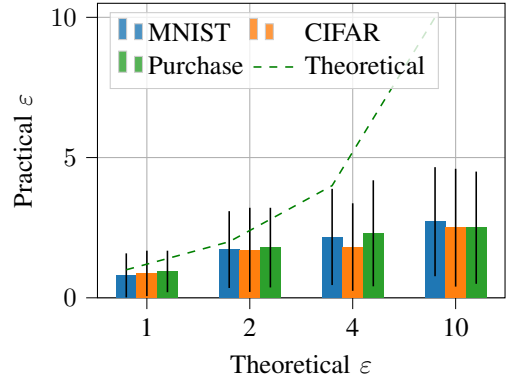
\begin{figure}[t]
    \centering
\begin{tikzpicture}

\definecolor{color0}{rgb}{0.12156862745098,0.466666666666667,0.705882352941177}
\definecolor{color1}{rgb}{1,0.498039215686275,0.0549019607843137}
\definecolor{color2}{rgb}{0.172549019607843,0.627450980392157,0.172549019607843}

\begin{axis}[
height=\figureheight,
width=\figurewidth,
legend cell align={left},
legend columns=2,
legend style={fill opacity=0.8, draw opacity=1, text opacity=1, at={(0.03,0.97)}, anchor=north west, draw=white!80!black},
tick align=outside,
tick pos=left,
x grid style={white!69.0196078431373!black},
xlabel={Theoretical \(\displaystyle  \varepsilon\)},
xmajorgrids,
xmin=-0.28, xmax=3.68,
xtick style={color=black},
xtick={0.2,1.2,2.2,3.2},
xticklabels={1,2,4,10},
y grid style={white!69.0196078431373!black},
ylabel={Practical \(\displaystyle  \varepsilon\)},
ymajorgrids,
ymin=0, ymax=10.5,
ytick style={color=black}
]
\draw[draw=none,fill=color0] (axis cs:-0.1,0) rectangle (axis cs:0.1,0.8);
\addlegendimage{ybar,ybar legend,draw=none,fill=color0};
\addlegendentry{MNIST}

\draw[draw=none,fill=color0] (axis cs:0.9,0) rectangle (axis cs:1.1,1.72);
\draw[draw=none,fill=color0] (axis cs:1.9,0) rectangle (axis cs:2.1,2.17);
\draw[draw=none,fill=color0] (axis cs:2.9,0) rectangle (axis cs:3.1,2.714);
\draw[draw=none,fill=color1] (axis cs:0.1,0) rectangle (axis cs:0.3,0.87);
\addlegendimage{ybar,ybar legend,draw=none,fill=color1};
\addlegendentry{CIFAR}

\draw[draw=none,fill=color1] (axis cs:1.1,0) rectangle (axis cs:1.3,1.71);
\draw[draw=none,fill=color1] (axis cs:2.1,0) rectangle (axis cs:2.3,1.81);
\draw[draw=none,fill=color1] (axis cs:3.1,0) rectangle (axis cs:3.3,2.5);
\draw[draw=none,fill=color2] (axis cs:0.3,0) rectangle (axis cs:0.5,0.94);
\addlegendimage{ybar,ybar legend,draw=none,fill=color2};
\addlegendentry{Purchase}

\draw[draw=none,fill=color2] (axis cs:1.3,0) rectangle (axis cs:1.5,1.79);
\draw[draw=none,fill=color2] (axis cs:2.3,0) rectangle (axis cs:2.5,2.3);
\draw[draw=none,fill=color2] (axis cs:3.3,0) rectangle (axis cs:3.5,2.5);
\path [draw=black, semithick]
(axis cs:0,0.015)
--(axis cs:0,1.585);

\path [draw=black, semithick]
(axis cs:1,0.351)
--(axis cs:1,3.089);

\path [draw=black, semithick]
(axis cs:2,0.455)
--(axis cs:2,3.885);

\path [draw=black, semithick]
(axis cs:3,0.77)
--(axis cs:3,4.658);

\path [draw=black, semithick]
(axis cs:0.2,0.0599999999999999)
--(axis cs:0.2,1.68);

\path [draw=black, semithick]
(axis cs:1.2,0.21)
--(axis cs:1.2,3.21);

\path [draw=black, semithick]
(axis cs:2.2,0.25)
--(axis cs:2.2,3.37);

\path [draw=black, semithick]
(axis cs:3.2,0.4)
--(axis cs:3.2,4.6);

\path [draw=black, semithick]
(axis cs:0.4,0.2)
--(axis cs:0.4,1.68);

\path [draw=black, semithick]
(axis cs:1.4,0.37)
--(axis cs:1.4,3.21);

\path [draw=black, semithick]
(axis cs:2.4,0.41)
--(axis cs:2.4,4.19);

\path [draw=black, semithick]
(axis cs:3.4,0.5)
--(axis cs:3.4,4.5);

\addplot [semithick, green!50!black, dashed]
table {%
0 1
1 2
2 4
3 10
};
\addlegendentry{Theoretical}
\end{axis}

\end{tikzpicture}
    \caption{\textbf{Adaptive malicious input attack}: the adversary changes the training dataset in each iteration. The adversary can achieve better results compared to prior attacks.}
    \label{fig:adap_sgd}
\end{figure}

\paragraph{Results}
Figure~\ref{fig:adap_sgd} summarizes  results for this setting.
Compared to the previous attack, the adversary now obtains a tighter lower bound on privacy. For example, on the Purchase dataset with $\varepsilon=2$, this adaptive poisoning attack can achieve a lower-bound $\varepsilon_{lower}=0.37$ compared to $0.25$ in the prior section. Having the ability to modify the training dataset at each iteration was the missing key to effectively exploit access to the intermediate model updates and  obtain tighter bounds. 


\subsection{Gradient attack}

\label{sec:attack5}
By taking a closer look at the DP-SGD formulation, we see that the analysis assumes the adversary is allowed to control not only the input examples $x$, but the gradient updates $\nabla_\theta \ell(f_\theta, x, y)$ themselves.
The reason this is allowed is because the clipping and noising is applied at the level of gradient updates, regardless of how they were obtained.
Thus, even though we intuitively know that the update vector was generated by taking the gradient of some function, the analysis does not make this assumption anywhere.
While this assumption might not be true in every setting, in federated learning~\cite{mcmahan2017communication,konevcny2016federated,shokri2015privacy} the participants can directly modify the gradient vectors, which can be a possible setting to deploy such an attack. Nevertheless, we are interested in evaluating this adversary primarily to understand what additional power this gives the adversary.

All prior attacks relied on measuring the model's loss on some input example to distinguish between the two datasets.
Unlike the previous attacks, the \algb{} this time will directly inspect the weights of the neural network.

\begin{protocol}
\centering
\begin{tabular}{lcrr}
\multicolumn{3}{c}{\textbf{Gradient Poisoning Adversary Game}} \\
\textbf{Adversary}\hspace{2em} & \hspace{3em} & \textbf{Model Trainer} \\
\cmidrule{1-3}
 & & $b \sample \{0, 1\}$ \\
\multicolumn{3}{c}{\sendmessageleft{top={$f_{\theta_{0}}$}}} \\
\multicolumn{2}{l}{$\mathcal{G}, \mathcal{G}' \sample \mathcal{A}(f_{\theta_i}) $ } & &
\rdelim\}{4}{*}[\rotatebox{90}{\hspace{-3em}Repeat $N$ Iterations}] \\
\multicolumn{3}{c}{\sendmessageright{top={$B^i = \{\mathcal{G}', \mathcal{G}\}$}}} \\
& & $B \gets \mathbb{B}(B_b^i)$ \\
\multicolumn{3}{r}{$B_{dp} \gets \{\text{clip}_p(x) + N \,\colon\,x \in B\}$} \\
\multicolumn{3}{r}{$\theta_{i+1} \gets \theta_{i} - \eta \sum_{x \in B_{dp}} x$} \\
\multicolumn{3}{c}{\sendmessageleft{top={$f_{\theta_{i+1}}$}}} \\
& \vdots & \\

\multicolumn{2}{l}{$s \gets \mathcal{B}(\{f_{\theta_i}\}, \mathcal{G})$} & & \\
\multicolumn{3}{c}{\sendmessageright{top={$s \in \{0,1\}$}}} \\
& & Check if $s=b$ \\
\end{tabular}
\caption{\textbf{Adversary game for gradient poisoning attack.} 
    \textbf{Round 2i. }The adversary chooses a collection of gradients $\mathcal{G},\mathcal{G}'$.
    \textbf{Round 2i+1. }The model trainer chooses a subset of one of these gradient updates,
    and updates the parameters using the clipped and noised gradients.
    \textbf{Round 2N. }The adversary predicts which dataset was used for training.}
    \label{fig:protocol5}
\end{protocol}

\paragraph{The \alga{}} (\cref{alg:gradient})
The \alga{} inserts a watermark into the model parameters. 
To minimize the effect of modifying these parameters, 
the adversary selects a set of model parameters
which leave the model's performance on training data largely unaffected.
The \alga{}
modifies model parameters whose gradients are smallest in magnitude: intuitively, these model parameters are not updated much during training so they are good candidates for being modified. 
In particular for the first $t_o$ iterations of  training, the adversary observes the model parameters' gradient and after $t_o$ iterations, it selects $2n$ parameters which have the smallest sum of gradient absolute values. \cref{alg:gradient} outlines this procedure.
\begin{crafter}[t]
\small
 \caption{Gradient attack}
 \begin{algorithmic}[1]
  \REQUIRE intermediate models up to step $T$ $f^1,\cdots,f^t$, clipping norm $C$, number of poison parameters $2n$, number of measurements $T_o$, training dataset $D$
  \STATE $M \gets \sum_{t=1}^{\min (T_o, T) -1} |f^{t+1} -f^{t}|  $
  \STATE $points \gets \text{select smallest 2n arguments from} M $
  \STATE $\nabla_{malicious} \gets \vec{0}$ 
  \STATE $s \gets 1$
  \FOR{$p$ in $points$}
  \STATE $\nabla_{malicious}[p] = s \frac{C}{\sqrt{2n}}$
  \STATE $s = -s $
  \ENDFOR
  \STATE 
  \STATE $\mathcal{G} \gets \{\}$
  \FOR{$(x,y) \in D$ }
  \STATE $\mathcal{G}.\text{insert}(\nabla l(f,x,y))$
  \ENDFOR
 \RETURN $\mathcal{G} , \mathcal{G} \cup \nabla_{malicious}$
 \end{algorithmic}
 \label{alg:gradient}
 \end{crafter}

\paragraph{Model training}
The model randomly selects to train on $D$ or $D'$. If the trainer selects $D'$, the trainer calls~\cref{alg:gradient} to obtain a malicious gradient. It also selects a batch from the training dataset and computes the corresponding private gradient update, then with probability $q$ it adds the malicious gradient from~\cref{alg:gradient}. The model trainer reveals the model parameters for all of the intermediate steps to the attacker.

\paragraph{The \algb{}} (\cref{alg:gradient_detection})
Given the collection of model weights $\{\theta_i\}_{i=1}^N$,
the \algb{} adversary will directly inspect the model weights to make its guess. Similar to the \alga{}, the adversary observes the gradient of the first $t_o$ iterations to find which $2n$ model parameters has the smallest overall absolute value. Since the \alga{} tries to increase the distance between these model parameters, to detect if such watermark exist in the model parameters, the \algb{} computes the \emph{distance} between the parameters, and then performs hypothesis testing to  detect the presence of a watermark.
The distance from the set of parameters  either  come from the gradient distribution ($Z$) plus the added Gaussian noise (i.e, null hypothesis), or  are watermarked which means they  come from the gradient distribution plus the added Gaussian noise plus the watermark: 
\begin{align*}
    H_{null} \,:\,\,  Z + \mathcal{N}(0, \sigma^2) \quad 
    H_{watermark} \,:\, Z + \mathcal{N}(\frac{C}{\sqrt{2n}}, \sigma^2) 
\end{align*}

Since the \alga{} chooses  the parameters with smallest changes, we assume the gradient of these points are zero. This lets us  simplify the problem to finding out if the distance between the parameters are coming from a Gaussian with mean zero or with mean equal to the sum of the distances between the watermarked parameters. Now, given the distance between the watermarked parameters, we can compute the likelihood of it coming from the watermarked gradient as:
\begin{align}
 p(Distance=d|watermark) =   \frac{1}{2\sqrt{\pi n \sigma^2}} \exp{- (\frac{(x-C)^2}{2 \sigma ^2})  }
\end{align}
where $2n$ is the number of selected parameters, $C$ and $\sigma$ are the clipping norm and the scale used to clip and noise gradients in DP-SGD,  and $d$ is the distance computed by~\cref{alg:gradient_detection}

\begin{distinguisher}[t]
\small
 \caption{Watermark detection}
 \begin{algorithmic}[1]
  \REQUIRE subsampling rate $q$, number of iterations $T$, clipping norm $C$, number of poison parameters $2n$,number of measurements $t_o$ , trained models $f_0,\cdots f_T$, threshold $\tau$
  \STATE $M \gets \vec{0}$
  \STATE $p_{watermark}  \gets 1$
  \FOR{$t=1$ to $T$}
  \IF {$t \leq t_o$}
  \STATE $M = M + |f_t - f_{t-1} |$
  \ELSE
  \STATE $points \gets \text{select smallest 2n arguments from} M $
  \STATE $\nabla_{malicious} \gets \vec{0}$ 
  \STATE $s \gets 1$
  \STATE $d \gets 0$
  \FOR{$p$ in $points$}
  \STATE $d = d +  s \cdot f_t-f_{t-1} [p]$
  \STATE $s = -s $
  \ENDFOR
  \STATE $p_{watermark}  = p_{watermark} \times p(d|watermark)$ 
  \ENDIF
  \ENDFOR
      \IF{ $p_{watermark} \geq \tau$}
  \RETURN $D'$
  \ENDIF
  \RETURN $D$
 \end{algorithmic}
 \label{alg:gradient_detection}
 \end{distinguisher}

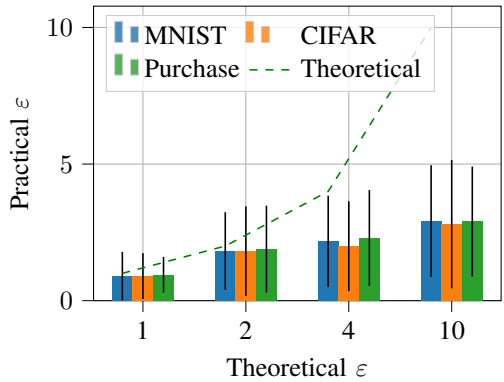
\begin{figure}[t]
    \centering
\begin{tikzpicture}

\definecolor{color0}{rgb}{0.12156862745098,0.466666666666667,0.705882352941177}
\definecolor{color1}{rgb}{1,0.498039215686275,0.0549019607843137}
\definecolor{color2}{rgb}{0.172549019607843,0.627450980392157,0.172549019607843}

\begin{axis}[
height=\figureheight,
width=\figurewidth,
legend cell align={left},
legend columns=2,
legend style={fill opacity=0.8, draw opacity=1, text opacity=1, at={(0.03,0.97)}, draw=white!80!black, anchor=north west,},
tick align=outside,
tick pos=left,
x grid style={white!69.0196078431373!black},
xlabel={Theoretical \(\displaystyle  \varepsilon\)},
xmajorgrids,
xmin=-0.28, xmax=3.68,
xtick style={color=black},
xtick={0.2,1.2,2.2,3.2},
xticklabels={1,2,4,10},
y grid style={white!69.0196078431373!black},
ylabel={Practical \(\displaystyle  \varepsilon\)},
ymajorgrids,
ymin=0, ymax=10.78,
ytick style={color=black}
]
\draw[draw=none,fill=color0] (axis cs:-0.1,0) rectangle (axis cs:0.1,0.9);
\addlegendimage{ybar,ybar legend,draw=none,fill=color0};
\addlegendentry{MNIST}

\draw[draw=none,fill=color0] (axis cs:0.9,0) rectangle (axis cs:1.1,1.82);
\draw[draw=none,fill=color0] (axis cs:1.9,0) rectangle (axis cs:2.1,2.17);
\draw[draw=none,fill=color0] (axis cs:2.9,0) rectangle (axis cs:3.1,2.914);
\draw[draw=none,fill=color1] (axis cs:0.1,0) rectangle (axis cs:0.3,0.9);
\addlegendimage{ybar,ybar legend,draw=none,fill=color1};
\addlegendentry{CIFAR}

\draw[draw=none,fill=color1] (axis cs:1.1,0) rectangle (axis cs:1.3,1.81);
\draw[draw=none,fill=color1] (axis cs:2.1,0) rectangle (axis cs:2.3,1.996);
\draw[draw=none,fill=color1] (axis cs:3.1,0) rectangle (axis cs:3.3,2.8);
\draw[draw=none,fill=color2] (axis cs:0.3,0) rectangle (axis cs:0.5,0.95);
\addlegendimage{ybar,ybar legend,draw=none,fill=color2};
\addlegendentry{Purchase}

\draw[draw=none,fill=color2] (axis cs:1.3,0) rectangle (axis cs:1.5,1.89);
\draw[draw=none,fill=color2] (axis cs:2.3,0) rectangle (axis cs:2.5,2.3);
\draw[draw=none,fill=color2] (axis cs:3.3,0) rectangle (axis cs:3.5,2.9);
\path [draw=black, semithick]
(axis cs:0,0.015)
--(axis cs:0,1.785);

\path [draw=black, semithick]
(axis cs:1,0.401)
--(axis cs:1,3.239);

\path [draw=black, semithick]
(axis cs:2,0.5)
--(axis cs:2,3.84);

\path [draw=black, semithick]
(axis cs:3,0.87)
--(axis cs:3,4.958);

\path [draw=black, semithick]
(axis cs:0.2,0.0599999999999999)
--(axis cs:0.2,1.74);

\path [draw=black, semithick]
(axis cs:1.2,0.17)
--(axis cs:1.2,3.45);

\path [draw=black, semithick]
(axis cs:2.2,0.35)
--(axis cs:2.2,3.642);

\path [draw=black, semithick]
(axis cs:3.2,0.45)
--(axis cs:3.2,5.15);

\path [draw=black, semithick]
(axis cs:0.4,0.3)
--(axis cs:0.4,1.6);

\path [draw=black, semithick]
(axis cs:1.4,0.3)
--(axis cs:1.4,3.48);

\path [draw=black, semithick]
(axis cs:2.4,0.55)
--(axis cs:2.4,4.05);

\path [draw=black, semithick]
(axis cs:3.4,0.89)
--(axis cs:3.4,4.91);

\addplot [semithick, green!50!black, dashed]
table {%
0 1
1 2
2 4
3 10
};
\addlegendentry{Theoretical}
\end{axis}

\end{tikzpicture}
    \caption{\textbf{Gradient attack:} the adversary directly insert a malicious gradient to training. The results are consistently better than the previous attack. }
    \label{fig:malgrad_sgd} 
\end{figure}

\paragraph{Results}
From Figure~\ref{fig:malgrad_sgd}, we derive that having direct access to the model gradients
further improves the strength of the adversary and establishes an improved lower bound.
For small values of $\varepsilon=1$, the attack can achieve empirical lower bound $\varepsilon=0.3$, this attack is almost tight,however, the gap increases as the value of epsilon increases.

\subsection{Malicious Datasets}
\label{sec:attack6}

In all of the previous attacks, we used a standard training 
dataset $D$, and  the attacker would create a malicious training instance to add to this dataset.
However, the DP-SGD analysis holds not just for a worst-case gradient from
some typical dataset (e.g., MNIST/CIFAR-10),
but also when the dataset itself is constructed to be worst-case.
It is unlikely this worst-case situation will ever occur in practice;
the purpose we study this set of assumptions is instead motivated by our
desire to establish tight privacy bounds on DP-SGD.

This attack corresponds identically to the protocol for the malicious gradient adversary.
Similarly, the \alga{} adversary is identical to the prior adversary, except that instead
of choosing $D$ to be some typical dataset we construct a new one (discussed next).
As in the prior attacks, we again use the intermediate-model \algb{} that has proven effective.

All that remains is for us to describe the method for choosing the dataset $D$.
Our goal in constructing this dataset is to come up with a dataset that will have minimal
unintended influence on the parameters of the neural network.

\paragraph{The \alga{}} (\cref{alg:mal_data})
To accomplish this, we will design the dataset $D$ so
that if $B = \mathbb{B}(D') \subset D$ then
training for one step on this mini-batch $\mathcal{S}(B)$ will 
minimally perturb the weights of the neural network.
We design ~\cref{alg:mal_data} to create the malicious dataset.
Informally, this algorithm proceeds as follows.
Using the initial random parameters $f_{\theta_0}$ of the neural network
(which are assumed to be revealed to the adversary in the DP-SGD analysis)
the first time the adversary is called 
we construct a dataset $D$ that the model already labels perfectly.
Because it is labeled perfectly, we will have
\[\nabla_{f_\theta}(\mathcal{L}(f_{\theta_0}, D)) \equiv 0 \,\,\,\,\,\,\, \forall D \subset \mathcal{D}\]
(otherwise it would be possible to construct a better dataset, leading to a contradiction)
and therefore training on any mini-batch that does not contain
$D' \backslash D$ will be an effective no-op---except
for the noise added through the Gaussian mechanism.

Unfortunately, this Gaussian noise the model adds will corrupt the model weights
and cause \emph{future} gradient updates to be non-zero.
To prevent that, we set the learning rate in the training algorithm to zero---esentially
ignoring the actual updates, and so $\theta_i \equiv \theta_0$.
This is allowable because DP-SGD guarantees that \emph{any} assignment of hyperparameters
will result in a private model---even if the choice of hyperparameters is pathalogical
and would never be used by a realistic adversary.

\begin{crafter}[H]
 \caption{Malicious dataset}
 \begin{algorithmic}[1]
  \REQUIRE initial model $f_0$, dataset size $n$, input space $\mathcal{D}$,  clipping norm $C$, selected params $2n$
  \STATE ${D}_{data} \gets $ randomly sample $n$ random instances from $\mathcal{D}$
  \STATE ${D}_{label} \gets f_0({D}_{data})$
  \STATE ${D} \gets ({D}_{data},{D}_{label})$
  \RETURN  Call \cref{alg:gradient} on ${D}$
 \end{algorithmic}
 \label{alg:mal_data}
 \end{crafter}

\paragraph{The \algb{}} (\cref{alg:gradient_detection}) Identical to prior.

\paragraph{Results}
Figure~\ref{fig:malgrad_sgd} shows how with a worst case dataset, the adversary can achieve a lower bound nearly tight with the upper bound provided by the analysis of DP-SGD. For example, when the theoretical bound for DP-SGD is $\varepsilon=4$, this last adversary  achieves a lower bound of $3.6$.

Compared to the previous adversaries, we clearly see that removing the ``intrinsic noise'' induced by the gradient updates from other examples tightens the lower bound significantly. We conclude that an adversary  taking advantage of all of the assumptions made in the analysis of DP-SGD is able to leak private information which is close to being maximal.
This suggests that the era of ``free privacy'' through better DP-SGD 
analysis has come to an end using existing assumptions.
If DP-SGD analysis were able to better capture the noise inherent to typical datasets, it might be possible to achieve substantially tighter guarantees---however, even formalizing  what this ``intrinsic noise'' means is nontrivial.




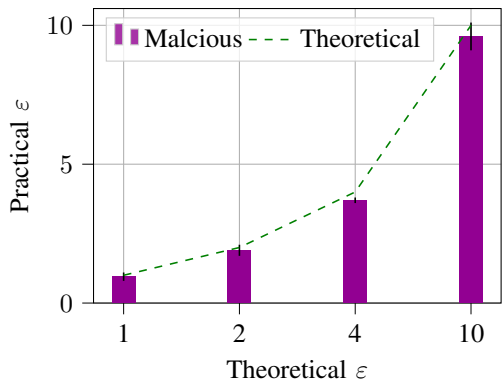
\begin{figure}
    \centering
\begin{tikzpicture}

\definecolor{color0}{rgb}{0.572549019607843,0.0,0.572549019607843}

\begin{axis}[
height=\figureheight,
width=\figurewidth,
legend cell align={left},
legend columns=3,
legend style={fill opacity=0.8, draw opacity=1, text opacity=1, at={(0.03,0.97)}, anchor=north west, draw=white!80!black},
tick align=outside,
tick pos=left,
x grid style={white!69.0196078431373!black},
xlabel={Theoretical \(\displaystyle  \varepsilon\)},
xmajorgrids,
xmin=-0.26, xmax=3.26,
xtick style={color=black},
xtick={0.0,1.0,2.0,3.0},
xticklabels={1,2,4,10},
y grid style={white!69.0196078431373!black},
ylabel={Practical \(\displaystyle  \varepsilon\)},
ymajorgrids,
ymin=0, ymax=10.605,
ytick style={color=black}
]
\draw[draw=none,fill=color0] (axis cs:-0.1,0) rectangle (axis cs:0.1,0.95);
\addlegendimage{ybar,ybar legend,draw=none,fill=color0};
\addlegendentry{Malcious}

\draw[draw=none,fill=color0] (axis cs:0.9,0) rectangle (axis cs:1.1,1.9);
\draw[draw=none,fill=color0] (axis cs:1.9,0) rectangle (axis cs:2.1,3.7);
\draw[draw=none,fill=color0] (axis cs:2.9,0) rectangle (axis cs:3.1,9.6);
\path [draw=black, semithick]
(axis cs:0,0.8)
--(axis cs:0,1.1);

\path [draw=black, semithick]
(axis cs:1,1.7)
--(axis cs:1,2.1);

\path [draw=black, semithick]
(axis cs:2,3.6)
--(axis cs:2,3.8);

\path [draw=black, semithick]
(axis cs:3,9.1)
--(axis cs:3,10.1);

\addplot [semithick, green!50!black, dashed]
table {%
0 1
1 2
2 4
3 10
};
\addlegendentry{Theoretical}
\end{axis}

\end{tikzpicture}
    \caption{\textbf{Malicious dataset attack}: the adversary creates a custom dataset to reduce the effect of other samples on the inserted watermark. This verifies the DP-SGD privacy is tight.  }
    \label{fig:grad_attack_fgd-maldata}
\end{figure}

%

\subsection{Theoretical Justifications}

Let us first consider a one dimensional learning task where the model space is $\mathcal{C}\subseteq \mathbb{R}$. 
Following the notation in the rest of the paper, private gradient descent essentially operates as: $\theta_{t+1}\leftarrow\theta_t - \eta \underbrace{\left(\nabla \mathcal{L}(f_{\theta_t};D)+Z_t\right)}_{v_t}$, where $Z_t$ is the Gaussian noise added at time step $t$. (We will ignore the details with clipping as it is orthogonal to the discussion here.) 
Clearly, for an adversary who observes $v_t$, the Gaussian noise $Z_t$ added 
to ensure R\'enyi differential privacy (RDP) at time step $t$ 
is both analytically and empirically tight. 
The analytical tightness follows from the tightness of Gaussian mechanism~\cite{mironov2017renyi}.
Hence, in the one dimensional case, we would expect our empirical lower bounds on privacy to be tight for one iteration.

Now, in higher dimensions, the question is how changing one sample $d\in D$ affects the gradient $\nabla \mathcal{L}(f_{\theta_t};D)$, which in turn affects the estimation of $v_t$ in various dimensions. First, we observe that by changing a data sample it is possible to affect only one coordinate of the gradient in the learning tasks we take up. Furthermore, since 
independent noise of the same scale is added
to each coordinate of $\nabla \mathcal{L}(f_{\theta_t};D)$, we can reduce the problem to the one dimensional case as above.

An additional step in DPSGD is subsampling for minibatch, which, according to~\cite{zhu2019poission,
wang2019subsampled}, has tight RDP guarantee.
Then, by nature of RDP, we can also tightly compose over iterations.
Finally, the conversion from RDP to differential privacy is also tight within constant factor.
Hence, the tightness of our results, especially those demonstrated by the gradient perturbation attacks, are natural.

\section{Related Work}
Our paper empirically studies differential privacy and differentially private stochastic gradient descent. Theoretical work producing upper bounds of privacy is extensive \cite{chaudhuri2011differentially,bassily2014private,song2013stochastic,abadi2016deep,mcmahan2017learning, wu2017bolt, iyengar2019towards, pichapati2019adaclip,feldman2020private,feldman2018privacy,abadi2016deep,bassily2014private,song2013stochastic,mironov2017renyi}.

Our work is closely related to other privacy attacks on machine learning models.
This work is mainly focused on developing attacks not for analysis purposes, but to demonstrate an attack \cite{carlini2019secret,fredrikson2015model}.
Jayaraman \emph{et al.} also consider the relationship between privacy upper bounds and lower bounds, but do not construct as strong lower bounds as we do \cite{jayaraman2019evaluating}.
Even more similar is Jagielski \emph{et al.} \cite{jagielski2020auditing},
who as we discussed earlier construct poisoning attacks to audit
differential privacy.
Our work is primarily different in that we construct lower bounds to measure the privacy assumptions, not just measure one set of configurations.
We additionally study much larger datasets (\cite{jagielski2020auditing} consider a two-class MNIST subset).

Our general approach is not specific to DP-SGD,
and should extend to other privacy-preserving training techniques.
For example, PATE \cite{papernot2018scalable} is an alternate technique to privately
train neural networks using an ensemble of teacher and student neural networks.
Instead of adversarially crafting datasets that will be fed to a training algorithm, we would craft datasets that would introduce different historgrams when the teacher is trained on the ensemble.

\section{Conclusion}
Our work provides a new way to investigate properties of
differentially private deep learning through the instantiation of games between hypothetical adversaries and the model trainer.
When applied to DP-SGD, this methodology allows us to evaluate the gap between the private information an attacker can leak (a lower bound) and what the privacy analysis establishes as being the maximum leak (an upper bound). 
%
Our results indicate that the current analysis of DP-SGD is tight (i.e., this gap is null) when the adversary
is given full assumed capabilities---however, when practical restrictions are placed
on the adversary there is a substantial gap between the upper and \fix{our} lower bounds.
This has two broad consequences.

\textbf{Consequences for theoretical research.}
Our work has immediate consequences for theoretical research on differentially private
deep learning.
For a time, given \emph{the same algorithm}, improvements to the analysis allowed
for researchers to obtain lower and lower values of $\varepsilon$.
Our results indicate this trend can continue no further.
We verify that the Moments Accountant \cite{bassily2014private,abadi2016deep} on DP-SGD as currently
implemented is tight.
As such, in order to provide better guarantees, either the DP-SGD algorithm will need
to be changed, or additional restrictions must be placed on the adversary.

Our work further provides guidance to the theoretical community for what additional assumptions might be most
fruitful to work with to obtain a better privacy guarantee.
For example, even if there were a way for DP-SGD to place assumptions on the
``naturalness'' of the training dataset $D$, it is unlikely to make a difference:
our attack that allows constructing a pathological dataset $D$ is only marginally
more effective than one that operates on  actual datasets like CIFAR10. 
On the contrary, our results indicate that restricting the adversary to only being allowed to modify an example
from the training dataset, instead of modifying a gradient update, is promising and could lead to more advantageous upper bounds.
\fix{While it is still possible that a stronger attack could establish a tighter bound, we believe that given the magnitude of this gap is unlikely to close entirely from below.}
%
%

\textbf{Consequences for applied research.}
Applied researchers have, for some time, chosen
``values of $\varepsilon$ that offer no meaningful theoretical guarantees'' \cite{carlini2019secret},
for example selecting $\varepsilon \gg 1$,
hoping that despite this ``the measured exposure [will be] negligible''~\cite{carlini2019secret}.
Our work refutes these claims in general.

When an adversary is assumed to have full capabilities made by DP-SGD,
they can succeed exactly as often as is expected given the analysis.
For example, in federated learning \cite{mcmahan2017communication}, adversaries \emph{are} allowed
to directly poison gradients and \emph{are} shown all intermediate model
updates. In the centralized setting, the situation is different: an adversary  assumed to have more realistic capabilities is \fix{currently} unable to succeed as often as the upper bound currently suggests.
\fix{However, conversely, it is always possible stronger adversaries could succeed more often.} 


\section*{Acknowledgements}
We are grateful to Jonathan Ullman, Andreas Terzis, and the anonymous reviewers for detailed comments on this paper. Milad Nasr is supported by a Google PhD Fellowship in Security
and Privacy.

\bibliographystyle{plain}
\bibliography{references}

\section*{Appendix A: Experimental Setup}
\label{sec:implementation}

We have implemented our algorithms in Objax, a machine learning library
that allows for efficiently training neural networks.
Because DP-SGD requires per-example gradients, instead of standard SGD
which just requires minibatch-averaged gradients, training models is
computationally expensive.
We leverage the JAX per-example parallel processing to speed up the training process.
(URL to open source repository blinded for review.)

\subsection{Datasets}\label{sec:appendix_dataset}
We run experiments on three datasets:

\paragraphb{MNIST}
We use the MNIST image dataset which  consists of 70,000 instances of $28\times28$  handwritten grayscale digit images and the task is to recognize the digits. MNIST splits the training data to 60,000 images for training and 10,000 for the testing phase.

\paragraphb{CIFAR}     We use CIFAR10, which  is a standard benchmark dataset consisting of $32\times 32$ RGB images. The learning task is to classify the images into 10 classes of different objects. This dataset is partitioned into 50,000 for training and 10,000 for testing. 

\paragraphb{Purchase}
The Purchase dataset is the shopping
records of several thousand online customers, extracted
during Kaggle’s ``acquire valued shopper'' challenge \cite{kaggle}. We used a processed version
of this dataset (courtesy of the authors of~\cite{shokri2017membership}). Each record
in the dataset is the shopping history of a single user. The
dataset contains 600 different products, and each user has a
binary record which indicates whether she has bought each of
the products. Records
are clustered into 10 classes based on the similarity of the
purchase. We use 18,000 records for training and 1,000 for  testing.  


\paragraphb{Hyper-parameters} For both CIFAR10 and MNIST, we used cross-entropy as our loss function. 
We used four layers convolutional neural network~\cite{goodfellow2016deep} for both MNIST and CIFAR10 datasets. For Purchase dataset, we used a three layers fully connected neural network. To train the models, we use differenatially private SGD~\cite{abadi2016deep} with 0.15 learning rate for MNIST and CIFAR and 1.1 for Purchase datasets, clipping factor of 1.0 for MNIST and CIFAR and 0.05 for Purchase dataset. We repeat MNIST, CIFAR and Purchase experiments 1000 each and  the malicious dataset 1,000,000 times. We evaluated each experiment using  noise multipliers (DPSGD parameter) of $0.5,0.6,0.9,1.0$ and picked the one which gives us the best practical epsilon. For MNIST, we get accuracy of $95\%$,$96.0\%$,$97\%$,$98.5\%$, on CIFAR we get $53\%, 55\%, 56\%, 59\%$, and on Purchase $37\%, 82\%, 88\%, 90\%$ for epsilons of $1,2,4,10$ respectively.

\begin{figure}
    \centering
\begin{tikzpicture}

\definecolor{color0}{rgb}{0.12156862745098,0.466666666666667,0.705882352941177}
\definecolor{color1}{rgb}{1,0.498039215686275,0.0549019607843137}
\definecolor{color2}{rgb}{0.172549019607843,0.627450980392157,0.172549019607843}
\definecolor{color3}{rgb}{0.572549019607843,0.0,0.572549019607843}

\begin{axis}[
height=\figureheight,
width=\figurewidth,
legend cell align={left},
legend style={fill opacity=0.8, draw opacity=1, text opacity=1, at={(0.03,0.97)}, anchor=north west, draw=white!80!black},
tick align=outside,
tick pos=left,
x grid style={white!69.0196078431373!black},
xmajorgrids,
xmin=-0.3, xmax=5.3,
xtick style={color=black},
y grid style={white!69.0196078431373!black},
ymajorgrids,
ymin=0, ymax=3.675,
ytick style={color=black},
xticklabels={,API,Blackbox,Whitebox,Adaptive,Gradient,Dataset},
x tick label style={rotate=45,anchor=east},
ylabel={measured privacy}
]
\draw[draw=none,fill=color0] (axis cs:-0.1,0) rectangle (axis cs:0.1,0.22);
\addlegendimage{ybar,ybar legend,draw=none,fill=color0};
\addlegendentry{MNIST}

\draw[draw=none,fill=color0] (axis cs:0.9,0) rectangle (axis cs:1.1,1.22);
\draw[draw=none,fill=color0] (axis cs:1.9,0) rectangle (axis cs:2.1,1.32);
\draw[draw=none,fill=color0] (axis cs:2.9,0) rectangle (axis cs:3.1,1.72);
\draw[draw=none,fill=color0] (axis cs:3.9,0) rectangle (axis cs:4.1,1.82);
\draw[draw=none,fill=color1] (axis cs:0.1,0) rectangle (axis cs:0.3,1.53);
\addlegendimage{ybar,ybar legend,draw=none,fill=color1};
\addlegendentry{CIFAR}

\draw[draw=none,fill=color1] (axis cs:1.1,0) rectangle (axis cs:1.3,1.51);
\draw[draw=none,fill=color1] (axis cs:2.1,0) rectangle (axis cs:2.3,1.61);
\draw[draw=none,fill=color1] (axis cs:3.1,0) rectangle (axis cs:3.3,1.71);
\draw[draw=none,fill=color1] (axis cs:4.1,0) rectangle (axis cs:4.3,1.81);
\draw[draw=none,fill=color2] (axis cs:0.3,0) rectangle (axis cs:0.5,1.62);
\addlegendimage{ybar,ybar legend,draw=none,fill=color2};
\addlegendentry{Purchase}

\draw[draw=none,fill=color2] (axis cs:1.3,0) rectangle (axis cs:1.5,1.59);
\draw[draw=none,fill=color2] (axis cs:2.3,0) rectangle (axis cs:2.5,1.59);
\draw[draw=none,fill=color2] (axis cs:3.3,0) rectangle (axis cs:3.5,1.79);
\draw[draw=none,fill=color2] (axis cs:4.3,0) rectangle (axis cs:4.5,1.89);
\draw[draw=none,fill=color3] (axis cs:4.85,0) rectangle (axis cs:5.15,1.98);
\addlegendimage{ybar,ybar legend,draw=none,fill=color3};
\addlegendentry{Malicious}

\path [draw=black, semithick]
(axis cs:0,0.01)
--(axis cs:0,0.43);

\path [draw=black, semithick]
(axis cs:1,0.221)
--(axis cs:1,2.219);

\path [draw=black, semithick]
(axis cs:2,0.2451)
--(axis cs:2,2.3949);

\path [draw=black, semithick]
(axis cs:3,0.351)
--(axis cs:3,3.089);

\path [draw=black, semithick]
(axis cs:4,0.401)
--(axis cs:4,3.239);

\path [draw=black, semithick]
(axis cs:0.2,0.13)
--(axis cs:0.2,2.93);

\path [draw=black, semithick]
(axis cs:1.2,0.2)
--(axis cs:1.2,2.82);

\path [draw=black, semithick]
(axis cs:2.2,0.2)
--(axis cs:2.2,3.02);

\path [draw=black, semithick]
(axis cs:3.2,0.21)
--(axis cs:3.2,3.21);

\path [draw=black, semithick]
(axis cs:4.2,0.25)
--(axis cs:4.2,3.37);

\path [draw=black, semithick]
(axis cs:0.4,0.1)
--(axis cs:0.4,3.14);

\path [draw=black, semithick]
(axis cs:1.4,0.28)
--(axis cs:1.4,2.9);

\path [draw=black, semithick]
(axis cs:2.4,0.25)
--(axis cs:2.4,2.93);

\path [draw=black, semithick]
(axis cs:3.4,0.37)
--(axis cs:3.4,3.21);

\path [draw=black, semithick]
(axis cs:4.4,0.45)
--(axis cs:4.4,3.33);

\path [draw=black, semithick]
(axis cs:5,1.9)
--(axis cs:5,2.06);

\addplot [semithick, red, dashed]
table {%
-0.3 2
5.3 2
};
\addlegendentry{Theoretical}
\end{axis}

\end{tikzpicture}
\hspace*{1cm}
\begin{tikzpicture}
\node  (A) {Practical};
\node [right=3cm of A] (B) {Theoretical};
\draw [->] (A) -- (B);
\end{tikzpicture}
    \caption{\textbf{Summary of our results}, plotting emperically measured $\varepsilon$
    when training a model with
    $\varepsilon=2$ differential privacy.
    The dashed red line corresponds to the certifiable upper bound.
    Each bar correspond to the privacy offered by
    increasingly powerful adversaries.
    In the most realistic setting, training with privacy offers much more
    empirically measured privacy.
    When we provide full attack capabilities,
    our lower bound shows that the DP-SGD upper bound
    is tight.
    }
    \label{fig:main_all}
\end{figure}
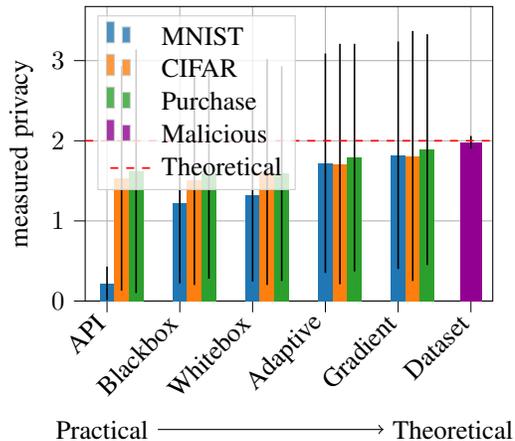

\section*{Appendix B: Summary of Results}

Figure~\ref{fig:main_all} shows an overview of the results in the paper. We target a specific theoretical epsilon for DPSGD for different datasets. Since all of the training datasets has between $10^4$ and $10^5$ instances, we choose $\delta=10^{-5}$ which means  we can leak one instance from the training dataset. 
This setting of $\delta$ is typical for DP-SGD \cite{abadi2016deep}.
The dataset attack can achieve a practical epsilon very close to the theoretical bounds. This result suggest that if an adversary has access to all of the assumptions in DPSGD then the theoretical analysis is tight and the researchers should focus more on relaxing some of the assumptions in the analysis. The gradient attack results are also close to the theoretical bounds which means the adversary does not need to modify the dataset. Going from having access to gradient to only input, we can see a noticeable drop in the empirical privacy parameters. When the adversary has access to intermediate step models (e.g., in federated learning) it can still achieve a close privacy parameters close to theoretical values. However, when the adversary does not maliciously modify the training dataset, gradients or input the practical epsilon is much lower than the theoretical ones. 

Overall, the results suggest the tightness of DP-SGD in the worst case,
but loseness of DP-SGD in the average case.

\section{Different Analyses of DP-SGD}
\begin{figure}
    \centering
\begin{tikzpicture}

\definecolor{color0}{rgb}{0.12156862745098,0.466666666666667,0.705882352941177}
\definecolor{color1}{rgb}{1,0.498039215686275,0.0549019607843137}

\begin{axis}[
height=\figureheight,
width=\figurewidth,
legend cell align={left},
legend columns=2,
legend style={fill opacity=0.8, draw opacity=1, text opacity=1, at={(0.03,0.97)}, anchor=north west, draw=white!80!black},
tick align=outside,
tick pos=left,
x grid style={white!69.0196078431373!black},
xlabel={\(\displaystyle  \sigma\)},
xmajorgrids,
xmin=0.515, xmax=2.385,
xtick style={color=black},
y grid style={white!69.0196078431373!black},
ylabel={ \(\displaystyle  \varepsilon\)},
ymajorgrids,
ymin=0.287478338710814, ymax=13.7872107613907,
ytick style={color=black}
]
\addplot [semithick, color0]
table {%
0.6 12.0436746440506
0.689473684210526 7.77021792519619
0.778947368421053 5.55617283984187
0.868421052631579 4.28073025032344
0.957894736842105 3.50889873537184
1.04736842105263 2.97682667017077
1.13684210526316 2.60098222067987
1.22631578947368 2.31642830610835
1.31578947368421 2.09209510075557
1.40526315789474 1.91004670556023
1.49473684210526 1.75898405731731
1.58421052631579 1.63137894373921
1.67368421052632 1.52200094658376
1.76315789473684 1.42709285785535
1.85263157894737 1.34388062307153
1.94210526315789 1.27026822914425
2.03157894736842 1.20464039274882
2.12105263157895 1.14573083202092
2.21052631578947 1.09253189237941
2.3 1.04423105895281
};
\addlegendentry{RDP~\cite{mironov2019r}}
\addplot [semithick, color1]
table {%
0.6 13.1735865603598
0.689473684210526 8.20029191447031
0.778947368421053 5.83862109778874
0.868421052631579 4.4960333605387
0.957894736842105 3.64046716067387
1.04736842105263 3.05117667282368
1.13684210526316 2.62199532947995
1.22631578947368 2.29610065837108
1.31578947368421 2.04052372482323
1.40526315789474 1.83489384635794
1.49473684210526 1.66597561623812
1.58421052631579 1.52480960546298
1.67368421052632 1.40512089504187
1.76315789473684 1.30238686726623
1.85263157894737 1.2132671427762
1.94210526315789 1.13524192641552
2.03157894736842 1.06637517447141
2.12105263157895 1.00515515465732
2.21052631578947 0.950384468176949
2.3 0.901102539741716
};
\addlegendentry{GDP~\cite{dong2019gaussian}}
\end{axis}

\end{tikzpicture}
\caption{Comparison between R\'enyi~\cite{mironov2019r} and Gaussian Differential Privacy~\cite{dong2019gaussian}}
    \label{fig:gdp_rdp}
\end{figure}
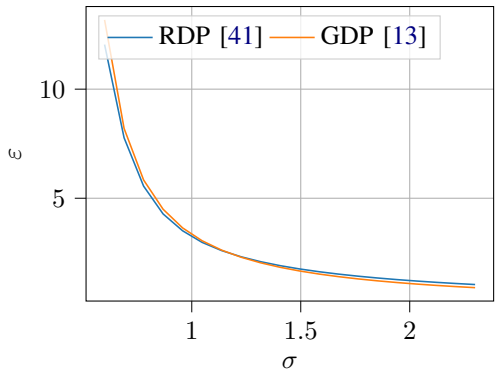

\fix{Abadi et al.~\cite{abadi2016deep} first introduced DP-SGD and the momentum accountant to analyze the privacy bounds. Since then, many works~\cite{dong2019gaussian,asoodeh2020better,wang2019subsampled,mironov2019r,koskela2020tight,balle2018privacy} improved the analysis of the DP-SGD to get better theoretical privacy bounds. As mentioned before, we used the R\'enyi Differential privacy (RDP)~\cite{mironov2019r} to compute the theoretical privacy bounds. Recently Dong et. al.~\cite{dong2019gaussian} introduced Gaussian differential privacy (GDP) to show a tight approach to compute the privacy bounds. However, both approaches result in similar privacy bounds. Using the hyperparameters for the MNIST dataset (sampling rate $\frac{256}{60000}$ and 60 epochs), we compute the final privacy costs using both R\'enyi and Gaussian differential privacy which is shown in the Figure~\ref{fig:gdp_rdp}. As we can see both results in comparable privacy bounds and are within our measurement error bounds. }
\end{document}